\documentclass[journal]{IEEEtran}
\usepackage{dsfont}

\usepackage{color}
\usepackage{setspace}
\usepackage{clrscode}
\usepackage{amsthm}
\usepackage{amsmath,bm}
\usepackage{pict2e}
\usepackage{amsmath}
\usepackage{amssymb}
\usepackage[pdftex]{graphicx}
\usepackage{bbold}
\usepackage[ruled,vlined]{algorithm2e}
\usepackage{caption}
\usepackage{subcaption}
\usepackage{cite}

\usepackage{mathabx}
\usepackage{stfloats}
\usepackage{multirow} 
\usepackage{float} 
\usepackage[ps2pdf,
bookmarks=false,
bookmarksnumbered=false, 
bookmarksopen=false, 
colorlinks=false]{}

\DeclareMathOperator*{\argmax}{argmax}
\DeclareMathOperator*{\argmin}{argmin}

\newcommand{\mV}{\mathcal{V}}
\newcommand{\mL}{\mathcal{L}}
\newcommand{\mT}{\mathcal{T}}
\newcommand{\mS}{\mathcal{S}}

\newcommand{\E}{\mathbb{E}}
\newcommand{\svec}{\textbf{s}}

\newcommand{\avec}{\textbf{a}}
\newcommand{\vvec}{\textbf{v}}
\newcommand{\vhvec}{\hat{\textbf{v}}}
\newcommand{\shvec}{\hat{\textbf{s}}}
\newcommand{\stvec}{\widetilde{\textbf{s}}}
\newcommand{\pvec}{\textbf{p}}
\newcommand{\ptvec}{\widetilde{\textbf{p}}}
\newcommand{\phvec}{\hat{\textbf{p}}}
\newcommand{\R}{\mathbb{R}}

\newcommand{\Pb}{\text{P}}

\newcommand{\dt}{\Delta t}
\newcommand{\xibm}{\bm{\xi}}

\allowdisplaybreaks[2]
\begin{document}
	\title{Learning-Based UAV Trajectory Optimization  with Collision Avoidance and Connectivity Constraints}
	
	\author{\IEEEauthorblockN{Xueyuan Wang and M. Cenk Gursoy}
			\thanks{The authors are with the Department of Electrical
				Engineering and Computer Science, Syracuse University, Syracuse, NY, 13244
				(e-mail: xwang173@syr.edu,  mcgursoy@syr.edu).}
	\thanks{The material in this paper will be presented in part at the IEEE International Conference on Communications (ICC) 2021.}	
		}

	
	\maketitle
	
	\begin{abstract}
		Unmanned aerial vehicles (UAVs) are expected to be an integral part of wireless networks, and determining collision-free trajectories for multiple UAVs while satisfying requirements of connectivity with ground base stations (GBSs) is a challenging task. In this paper, we first reformulate the multi-UAV trajectory optimization problem with collision avoidance and wireless connectivity constraints as a sequential decision making problem in the discrete time domain. We, then, propose a decentralized deep reinforcement learning approach to solve the problem. More specifically, a value network is developed to encode the expected time to destination given the agent's joint state (including the agent's information, the nearby agents' observable information, and the locations of the nearby GBSs). A signal-to-interference-plus-noise ratio (SINR)-prediction neural network is also designed, using accumulated SINR measurements obtained when interacting with the cellular network, to map the GBSs' locations into the SINR levels in order to predict the UAV's SINR.  Numerical results show that with the value network and SINR-prediction network, real-time navigation for multi-UAVs can be efficiently performed in various environments with high success rate.
	\end{abstract}

	\thispagestyle{empty}

	\section{Introduction}
	Unmanned aerial vehicles (UAVs), also commonly known as drones, are aircrafts piloted by remote control or embedded computer programs without human onboard \cite{UAV_survey_YZeng}. Recently,  UAVs have found numerous applications, such as aerial inspection, photography, precision agriculture, traffic control, search and rescue, package delivery, and telecommunications.
	Based on the roles of UAVs, the following two scenarios are considered to integrate the UAVs into cellular networks: 1) UAV-assisted cellular networks, in which UAVs can be deployed as aerial base stations (BSs) to support wireless connectivity and improve the performance of cellular networks \cite{UAV_CLiu}; 2) cellular-connected UAV networks, in which the UAVs are regarded as aerial user equipments (UEs) that need to be supported by the ground communication infrastructure \cite{UAV_cellular_MAzari_journal}.
	As aerial UEs, the UAVs need efficient trajectories and also should keep connected with ground base stations (GBSs) during their flights. Therefore, the trajectory of cellular-connected UAVs need to be carefully designed to meet their mission specifications, while at the same time ensuring that the communication requirements are satisfactorily met.

	Trajectory optimization for cellular-connected UAVs has been investigated in the literature,  in which a UAV has a  mission of flying between a pair of given initial and final locations.
	The authors in \cite{uavtraj_SZhang,uavtraj_EBulut,uavtraj_BKhamidehi}  addressed the trajectory optimization problem with the goal to minimize the UAV's mission completion time.
	Particularly, the authors in \cite{uavtraj_SZhang} considered how to determine the optimal path for the UAV, subject to a quality of connectivity constraint in the GBS-to-UAV link specified by a minimum receive signal-to-noise ratio target. Techniques from graph theory and convex optimization were used to find the trajectory solution. In \cite{uavtraj_EBulut}, the UAV is required to find a path during which it does not lose its cellular connection to one of the GBSs in the area by more than a given time period. Dynamic programming based approximate solution was proposed in this paper.
	The authors in \cite{uavtraj_BKhamidehi} studied the trajectory optimization problem for a UAV with two different criteria for the connectivity constraint: 1) the maximum continuous time duration that the UAV is out of the coverage of the GBSs is limited to a given threshold; 2) the total time periods that the UAV is not covered by the GBSs is restricted.	A double Q-learning method is proposed to solve the problem.
 	Moreover,	authors in \cite{uavtraj_YZeng} formulated a UAV trajectory optimization problem to minimize the weighted sum of its mission completion time and	expected communication outage duration. A dueling double deep Q network with multi-step learning algorithm is proposed. A simultaneous navigation and radio mapping framework was also proposed to improve the performance.
	 Additionally, an interference-aware path planning	scheme for a network of cellular-connected UAVs was proposed in \cite{uavtraj_UChallita}.  In particular, each UAV aims to achieve	a tradeoff between maximizing energy efficiency and minimizing both wireless latency and the interference caused on the ground network along its path.  A deep reinforcement learning algorithm, based on echo state network cells, was developed to solve the problem. In addition, trajectory design for cellular-connected UAVs has also been extensively investigated in  \cite{uavtraj_SZhang2, uavtraj_YZeng2,uavtraj_YZeng3, uavtraj_SDe, uavtraj_Bhamidehi,uavtraj_yang2019connectivity,uavtraj_zhang2020radio,uavtraj_mu2020non}.
	 However, none of the prior works considered multi-UAV networks, which is common in practice, along with collision avoidance constraints.
	
	 In scenarios involving multiple UAVs or more generally multiple autonomous systems, a fundamental challenge is to safely control the interactions with other dynamic agents in the environment.  Specifically, it is important for the autonomous devices (e.g., robots and drones) to navigate in an environment with or without obstacles, and stay free of collisions with each other and the obstacles,  based on local observations of the environment. Finding solutions to this problem is challenging, since one robot's action is based on others' motions (intents) and policies which are in general unknown, and, furthermore, explicit	communication of such hidden quantities is often impractical	due to physical limitations. Earlier works have largely leveraged well-engineered interaction models to enhance the social awareness in robot navigation, e.g. \cite{rvo_JBerg} and \cite{van2011reciprocal}, where the same policy is applied to all agents. The key challenge for these models is that they heavily rely on hand-crafted functions and cannot generalize well to various scenarios for crowd-like cooperation. As an alternative, reinforcement learning frameworks have been used to train computationally efficient policies that	implicitly encode the interactions and cooperation among	agents. Recent works, e.g., \cite{chen2016decentralized,DBLP_ChenELH17,CA_MEverett,everett2020collision}, have shown the power of deep reinforcement learning techniques to learn socially cooperative policies.
	
	 Different approaches for the collision avoidance of multiple UAVs have also been developed in the literature. For instance, a rolling horizon approach using dynamic programming  was used to solve the problem in a multi-agent cooperative system in \cite{UAV_CA_RBeard}. A neuro-dynamic programming algorithm is proposed in \cite{UAV_CA_DBauso} for multi-UAV cooperative path planning.
	 A mixed integer linear programming method is used in \cite{song2016rolling}. Partially observable Markov decision process based methods are applied in \cite{wolf2011aircraft,temizer2010collision,bai2012unmanned} for UAV collision avoidance.	
	  In addition, authors in \cite{UAV_CA_YLin} used reachable sets to represent the collection of possible trajectories of the obstacle aircraft. Once a collision is detected, a sampling-based method is used to generate a collision avoidance path for the UAV. The UAV in this paper was able to learn the position, velocity and receive other data from the obstacle aircraft.
	  In \cite{yu2017collision}, predictive state space was utilized to present the waypoints of the UAVs, with which initial collision-free trajectories are generated and then improved by a rolling optimization algorithm to minimize the trajectory length. 	
	  However, considering collision avoidance in multi-UAV navigation with wireless communication requirements and addressing these challenges via deep reinforcement learning methods have not been adequately explored yet.

	
	Motivated by these facts, we propose a decentralized deep reinforcement learning algorithm as a solution to the multi-UAV trajectory optimization problem  with collision avoidance and wireless connectivity constraints. The contributions of the paper are listed as follows:
	\begin{itemize}
		\item  We study multi-UAV trajectory optimization under realistic constraints, e.g., collision avoidance, wireless connectivity, and kinematic constraints, while also taking into account antenna patterns and interference levels.  Since it is difficult to address this problem with standard optimization techniques (especially in a decentralized setting), we further reformulate it as a sequential decision making problem in the discrete time domain.
		\item We develop a decentralized deep reinforcement learning algorithm to learn the action policy for each UAV. More specifically, we optimize the value function of the Markov decision process (MDP) transformed from the formulated problem.  Due to the high dimension and the continuity of the state space and action space, we design a value neural network to approximate the value function, and to encode the expected time to destination given the agent's joint state (including the agent's information, the nearby agents' observable information, and the locations of the nearby GBSs).
		\item Due to the fact that the UAVs do not communicate in the considered network, uncertainty exists in the UAVs' unobservable intents, which is critical for multi-UAV navigation problems. To address this uncertainty, we employ a  velocity-filter approach to estimate the UAVs' intentions.
		\item We further design a signal-to-interference-plus-noise ratio (SINR)-prediction neural network to assist the value network to encode the interaction between the UAVs and the cellular network. Particularly, using accumulated SINR measurements obtained when interacting with the cellular network, the SINR-prediction network maps the nearby GBSs' locations into the SINR levels in order to predict the UAV's SINR.
		\item We delineate the initialization, refining, and training steps of the algorithm and describe the real-time navigation process. We extensively evaluate the proposed decentralized deep reinforcement learning  algorithm.  We demonstrate that with the introduction of the SINR-prediction network, testing environment is not restricted to be the same as the training environment. Furthermore, we show that real-time decentralized navigation of multiple UAVs can be efficiently performed with high success rate in various environments, e.g., environments with different antenna patterns, environments with obstacles or no-fly zones.
	\end{itemize}
	
	The remainder of the paper is organized as follows: System model is introduced in Section II. Section III describes the multi-UAV trajectory optimization problem, including the considered constraints.
	Section IV focuses on the reinforcement learning framework for solving the proposed problem, and the approaches used to tackle the uncertainty in the environment.
	The decentralized deep reinforcement learning algorithm is presented in Section V in detail.
	In Section VI, numerical and simulation results are provided to evaluate the performance of the proposed algorithm.
	Finally, concluding remarks are given in Section VII.


	\begin{figure}		
		\centering
		\includegraphics[width=0.45\textwidth]{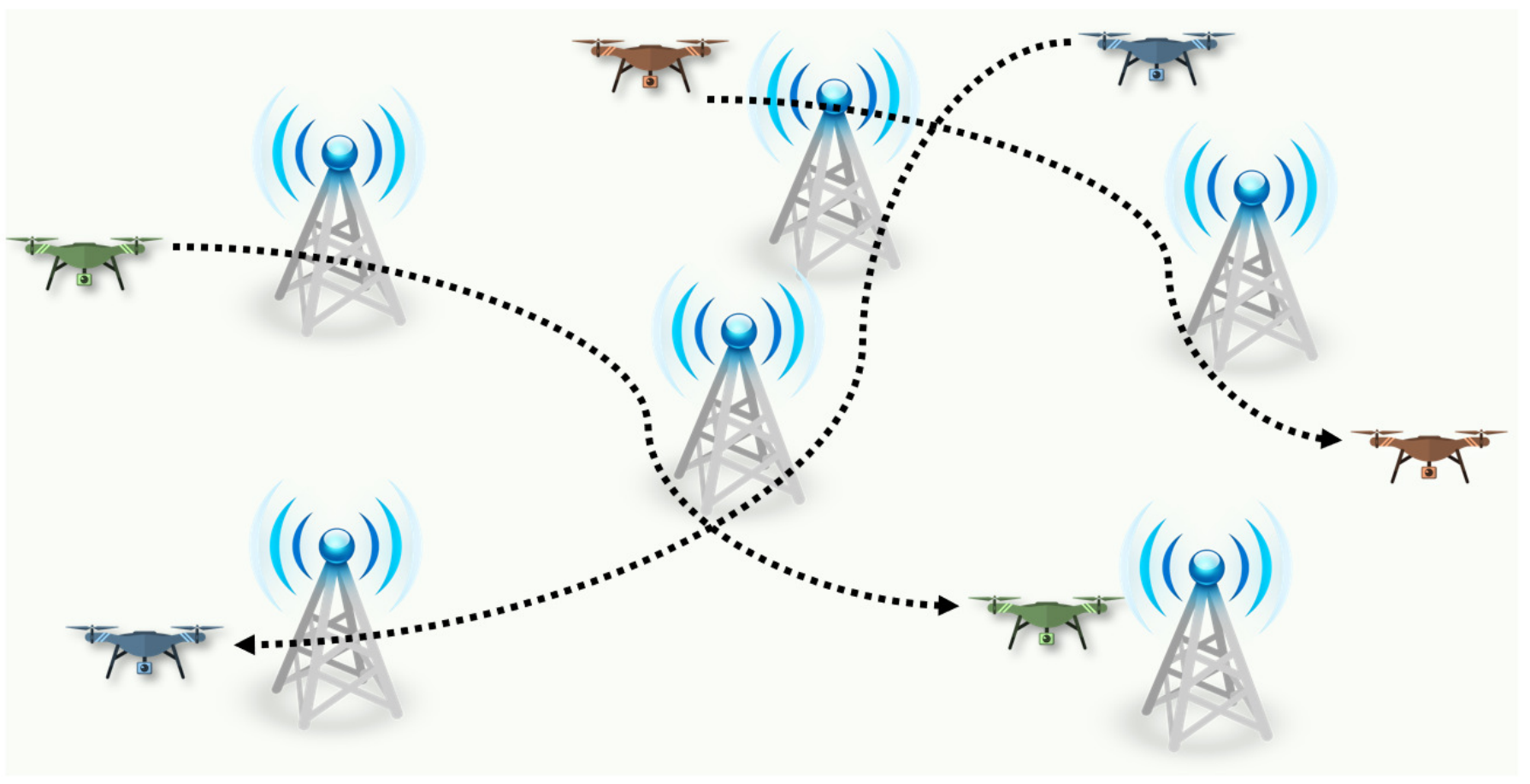}	
		\caption{\small An illustration of multi-UAV multi-GBS cellular networks.  \normalsize}
		\label{Fig:system_model}
	\end{figure}
	\section{System Model}
	In this section, we introduce the system model of the multi-UAV and multi-GBS cellular networks in detail.
	Note that in this section, unless specified otherwise, we remove the time index e.g., in the position vector $\pvec(t) \rightarrow \pvec$, and the index for UAVs or GBSs, e.g., $\pvec_i \rightarrow \pvec$.
	\subsection{Deployment}
	We consider multi-UAV multi-GBS cellular networks as displayed in Fig. \ref{Fig:system_model}, in which $J$ UAVs, with potentially different missions, need to fly from  starting locations to destinations over an area containing $K$ GBSs. Without loss of generality, we assume that the area of interest is a cubic volume, which can be specified by $C: \mathcal{X}\times \mathcal{Y} \times \mathcal{Z}$ and $\mathcal{X}\triangleq [x_{\min}, x_{\max}]$, $\mathcal{Y}\triangleq [y_{\min}, y_{\max}]$, and $\mathcal{Z}\triangleq [z_{\min}, z_{\max}]$.  Let $\pvec = [p_x, p_y, H_V]$ denote the 3D position of the UAV, where $H_V$ is the altitude of the UAVs which is assumed to be fixed for all UAVs. $\pvec^S =[p_{sx},p_{sy},H_V] \in \R^3$ and  $\pvec^D =[p_{gx}, p_{gy}, H_V] \in \R^3$ are used to denote the coordinates of the starting points and destinations.
	
	Each UAV's state is composed of an observable information vector and an unobservable (hidden) information vector, $\svec = [\svec^o, \svec^h]$, where the observable state can be observed by other UAVs, while the unobservable state can not. In the global frame, observable state includes the UAV's position, velocity $\vvec = [v_x,v_y]$, and radius $r$, i.e., $\svec^o = [\pvec, \vvec, r] \in \R^6$. The unobservable state consists of the destination $\pvec^D$, maximum speed $v_{\max}$, and orientation $\phi$, i.e., $\svec^h =[\pvec^D, v_{\max},\phi] \in \R^5$. It is worth noting that the UAVs do not communicate with other UAVs. Hence, we address a more challenging non-communicating scenario.
	
	In this cellular network, there are $K$ GBSs providing wireless  coverage  simultaneously.  The $k^{th}$ GBS has transmit power $P_{B_k}$, and it  is located at position $\pvec_{B_k} = [p_{x_{B_k}}, p_{y_{B_k}}, H_{B}]$, where $H_B$ is the height of the GBS and is assumed to be the same for all GBSs.

	\begin{figure}		
		\centering
		\includegraphics[width=0.3\textwidth]{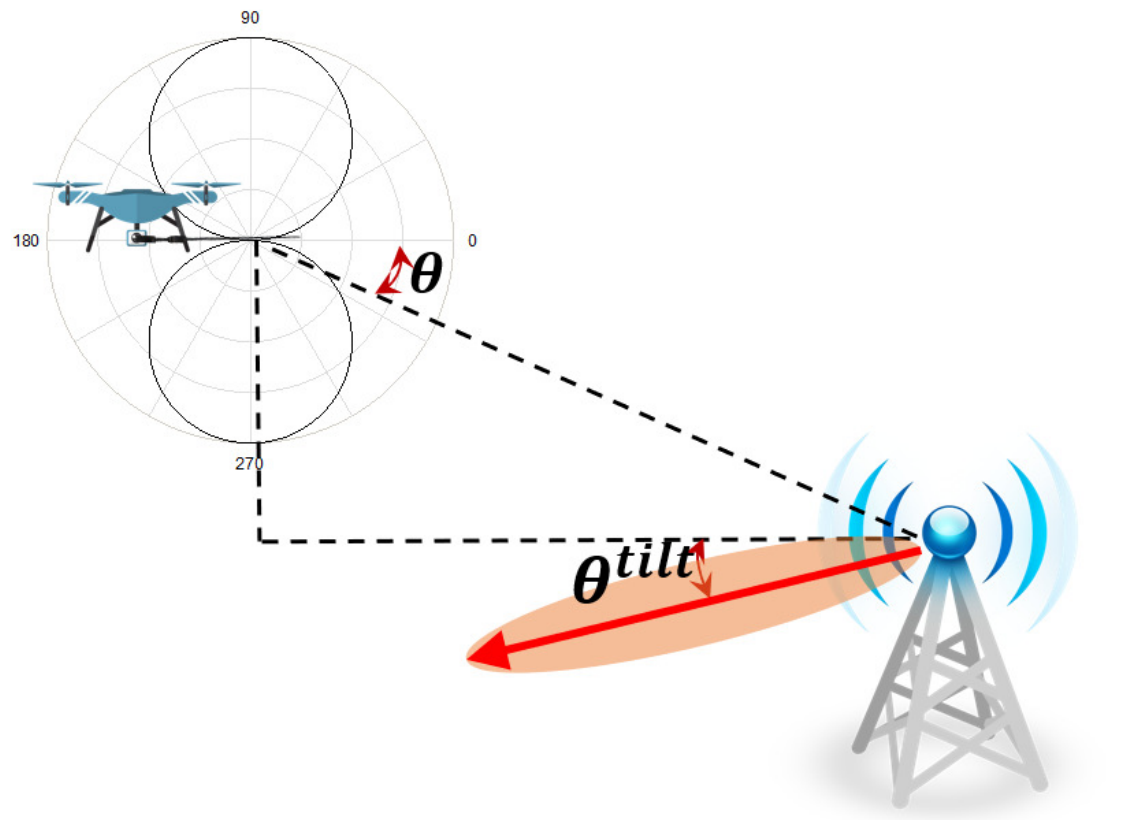}	
		\caption{\small Illustrations of the antenna patterns of the UAVs and the GBSs.  \normalsize}
		\label{Fig:antenna_pattern}
	\end{figure}
	\subsection{Antenna Configuration}
	The GBSs and the UAVs are equipped with directional antennas with fixed radiation patterns, which are shown in Fig. \ref{Fig:antenna_pattern}.
	\subsubsection{GBS}
	We assume that the antenna elements of the GBSs are only directional along the vertical dimension but omni-directional horizontally \cite{UAV_survey_YZeng}.
	Along the vertical dimension, the signal is usually downtilted toward the ground to cover the ground users and suppress the intercell interference.  Therefore, the antenna gain can be expressed as \cite{3GPP_36814}
	\begin{align}
	G_B(d) &= G_h + G_v  \text{ (dB)} \notag \\
	& = 10 ^{- \min\left(-1.2 \left(\frac{\arctan(\frac{H_B -H_V}{d})-\theta^{tilt}}{\theta^{3dB}}\right)^2, \frac{G_m}{10}\right)}
	\end{align}
	where
	\begin{align}
		&G_h = 0 \text{ (dB)} \\
		&G_v(d) 
		=  - \min \left(12 \left(\frac{\arctan(\frac{H_B -H_V}{d})-\theta^{tilt}}{\theta^{3dB}}\right)^2, G_m \right)  \text{ (dB)}.		
	\end{align}
	Above,  $G_m$ is the maximum attenuation of the antennas, $d$ is the horizontal distance between the UAV and the GBS, $\theta^{tilt}$ and $\theta^{3dB}$ represent electrical antenna downtilting angle and the vertical 3dB beamwidth of the antennas at the GBSs.
	\subsubsection{UAV}
	The UAVs are assumed to be equipped with a receiver with a horizontally oriented antenna, and a simple analytical approximation for antenna gain provided by UAVs can be expressed as \cite{UAV_Antenna_JChen}
	\begin{align}
	G_V (d) = \sin(\theta) = \frac{H_V- H_B}{\sqrt{d^2 + (H_V -H_B)^2}}
	\end{align}
	where $\theta$ is elevation angle between the UAV and GBS, $H_V$ is the UAV altitude, and $H_B$ is the height of the GBS.
	
	\subsection{Path Loss}
	 We assume that  the path loss can be expressed as
	\begin{align}
		L(d) =  \left(d^2 + (H_B-H_V)^2 \right) ^{\alpha/2}
	\end{align}
	where $\alpha$ is the path loss exponent.
	
	\subsection{SINR and Connectivity}
	The UAVs receive signals from all GBSs, among one of which is the serving BS, and others contribute to the interference. The received signal from the $k^{th}$ GBS to the $i^{th}$ UAV can be expressed as $P_k G_{B_k}(d_{ik}) G_{V_i}(d_{ik}) L^{-1}(d_{ik})$. The experienced SINR at the $i^{th}$ UAV if it is associated with the $k^{th}$ GBS an be expressed as
	\begin{align}
	\label{Eq:SINR}
	\mS_{r_{i,k}} \triangleq \frac{P_k G_{B_k}(d_{ik}) G_{V_i}(d_{ik}) L^{-1}(d_{ik}) }{\mathcal{N}_s  + \sum_{k' \neq k} P_{k'} G_{B_{k'}}(d_{i{k'}}) G_{V_i}(d_{i{k'}}) L^{-1}(d_{i{k'}}) }
	\end{align}
	where $\mathcal{N}_s $ is the noise power.
	If the experienced SINR at a UAV is larger than a threshold $\mT_{s}$, then the UAV is regarded as connected with the cellular network, and disconnected otherwise.
	
	\subsection{SINR Measurement}	
	Along the path to destination, UAVs interact with the cellular network, measure the raw signal from GBSs, and obtain the instantaneous SINR $\mS'_{r}([\pvec, \svec_{B}]; h)$, where $h$ includes the random small-scale fading coefficients with all GBSs, and $\pvec$ and $\svec_{B} = [\pvec_{B_k}, \forall k]$ are the position of the UAV and positions of all GBSs, respectively. These measurements can be obtained by leveraging the existing soft handover mechanisms with continuous reference signal received power (RSRP) and reference signal received quality (RSRQ) \cite{uavtraj_YZeng}.
	At each time $t$, over a very short time interval,  during which the agents' locations can be approximately considered to be unchanged, it is assumed that the UAV performs $N_m$ SINR measurements (each of which takes milliseconds). Then the empirical SINR can be obtained as
	\begin{align}
	\widehat{\mS}_{r_{(t)}} = \frac{1}{N_m} \sum_{n=1}^{N_m} \mS'_{r_{(t)}}([\pvec(t), \svec_{B}]; h_{(t),n}).
	\end{align}
	To average over the randomness arising from small-scale fading, we can consider large $N_m$ and have $\lim\limits_{N_m \rightarrow \infty}\widehat{\mS}_{r_{(t)}} = \mS_{r_{(t)}}$ by applying the law of large numbers.  Therefore, as long as the UAV performs signal measurements sufficiently frequently so that $N_m \gg 1$, $\mS_{r_{(t)}}$ can be evaluated by its empirical value $\widehat{\mS}_{r_{(t)}}$.

	\section{Multi-UAV Trajectory Optimization}
	In this section, we first introduce the constraints and then formulate the multi-UAV trajectory optimization problem.
	\subsection{Constraints}
	
		\subsubsection{Collision Avoidance}
		Collision avoidance is central to many autonomous systems. During flight, the UAVs should not collide with others, which means that the distance between two agents should be larger than their radius all the time, i.e.,
		\begin{align}
		||\pvec_i(t) - \pvec_j(t)  ||_2 > r_i+r_j \quad \forall j\neq i, \forall t
		\end{align}
		where $\pvec_i(t)$ is the location of the $i^{th}$ UAV at time $t$, and $r_i$ is its radius. Note that this radius can also include a buffer zone in which no other UAV should be present.
	
		\subsubsection{Wireless Connectivity Constraint}
		To support the command and control and also data flows, UAVs have to maintain a reliable communication link to the GBSs. To achieve this goal, we consider the  connectivity constraint for the UAVs, i.e., the maximum continuous time duration that the UAV is disconnected should not be longer than $\mT_t$ time units. The maximum continuous disconnected time duration can be mathematically expressed as
		\begin{align}
			T_O^{\max} = \max_{t \in [0,T]} t - T_L(t)
		\end{align}
		where $T$ is the total travel time, and  $T_L(t)$ is the last time that the UAV is connected with the cellular network  before time $t$, i.e.,
		\begin{align}
		   T_L(t) = \max  &\qquad  \tau \\
		   \text{s.t. }  &\quad  \tau \in [0,t]  \notag\\
		  				 &\quad \mS_{r} (\tau)\geq \mT_{s}. \notag
		\end{align}
		Therefore, the connectivity constraint can be written as
		\begin{align}
			\left(\max_{t \in [0,T]} t - T_L(t) \right) \leq \mT_t.
		\end{align}

		\subsubsection{Initial and Final Locations}
		Each UAV starts its mission from a given initial location and completes its flight at  a given destination, i.e., $\pvec(0) = \pvec^S$ and $\pvec(T) = \pvec^{D}$.
		
		\subsubsection{Kinematic Constraint}
		Kinematic constraints need to be considered for operating UAVs. We impose the speed and rotational constraints as follows:
		\begin{align}
		    &\label{Eq:kitc_const} \vvec(t) = [v_s(t), \phi(t)] \\
			&\label{Eq:kitc_const_v}\text{Speed limit: } v_s(t) \leq v_{\max} \\
			&\label{Eq:kitc_const_angl}\text{Rotation limit: } |\phi_(t)- \phi (t-\Delta t)| \leq \Delta t \cdot \mT_r
		\end{align}
		where $\vvec(t)$, $v_s(t)$ and $\phi(t)$ are the UAV's velocity, speed and orientation at time $t$. $v_{\max}$ is the maximum of speed the UAV, and $\mT_r$ is the maximum angle that a UAV can rotate in unit time period. This constraint limits the direction that a UAV can travel at a given time.
		
		\subsubsection{Association Constraint}
	    Each UAV is associated with one GBS at a time, and the associated GBS is denoted by $a(t) \in \{1, ..., K  \}$.
		
	
	\subsection{Problem Formulation in Continuous Time Domain}
	The goal of this work is to find trajectories for all UAVs in the network such that the travel/flight time of each UAV between the initial and final locations is minimized, while the constraints are satisfied. In the considered decentralized setting, the trajectory optimization problem for the $i^{th}$ UAV can be formulated as	
	\begin{align}
	(\Pb 0):\argmin_{ \{\pvec_i(t), a_i(t), \forall t\}}& \qquad T_i  \notag \\
	s.t.         \hspace{0.4in}  & \hspace{-0.3in}  ||\pvec_i(t) - \pvec_j(t)  ||_2 > r_i+r_j, \forall j \neq i, \forall t\tag{P0.a} \\
	&\hspace{-0.3in}  \max_{t \in [0,T]} t - T_L(t) \leq \mT_t \tag{P0.b} \\
	&\hspace{-0.3in}  \pvec_i(0) = \pvec_i^S, \pvec_i(T_i) = \pvec_i^{D} \tag{P0.c} \\
	&\hspace{-0.3in}  v_{s_i}(t) \leq v_{\max_i}, \forall t\tag{P0.d}\\
	&\hspace{-0.3in}  |\phi_i(t)- \phi_i(t-\Delta t)| \leq \Delta t \cdot \mT_r, \forall t \tag{P0.e}\\
	&\hspace{-0.3in}  a_i(t) \in \{1, ..., K  \}, \forall t 	\tag{P0.f}
	\end{align}
	
	\subsection{Problem Formulation in Discrete Time Domain}
	Since the UAV is not permitted to be disconnected continuously for more than $\mT_t$ time units, it is sufficient to consider $\Delta t= \mT_t/n_t $ as one time step and address the problem every $n_t$ time steps. If, at these specific time instances, the experienced SINRs at all  UAVs are higher than $\mT_s$, we can guarantee that the  connectivity constraint is satisfied. Now, the  optimization problem can be represented in discrete time domain as follows:
	\begin{align}
	(\Pb 1):\argmin_{ \{\pvec_{i,t}, a_{i,t}, \forall t\}}  & \qquad T_i \notag  \\
	s.t.         \quad  & ||\pvec_{i,t} - \pvec_{j,t}  ||_2 > r_i+r_j, \forall j \neq i, \forall t \tag{P1.a}\\
	& \mS_{r_{i,t}} \geq \mT_s, \text{ if } t \mid n_t \tag{P1.b}\\
	& \pvec_{i,0} = \pvec_i^S, \pvec_{i,T_i} = \pvec_i^{D}, \forall i \tag{P1.c}\\
	& v_{s_{i,t}} \leq v_{\max_i}, \forall t \tag{P1.d}\\
	& |\phi_{i,t}- \phi_{i,t-1}| \leq \Delta t \cdot \mT_r, \forall t \tag{P1.e}\\
	& a_{i,t} \in \{1, ..., K  \}, \forall t	\tag{P1.f}
	\end{align}
	where the integer-valued discrete time index $t$ indicates time increments by $\Delta t$, and $t \mid n_t$ signifies that $t$ is divisible by $n_t$.
	
	It is obvious that the optimal cell association policy can be  obtained as $	a_{i,t}^* = \argmax_{k \in \{1,..., K\}}  S_{r_{i,t,k}}$.
	Then (P1) reduces to
	\begin{align}
	(\Pb 2):\argmin_{ \{\pvec_{i,t},\forall t \} } &\qquad  T_i \\
	s.t.         &\quad (\Pb 1.a)-(\Pb 1.e).  \notag
	\end{align}
	
	The non-communicating multi-agent navigation task can be formulated as a sequential decision making problem in a reinforcement learning framework \cite{chen2016decentralized}. 	
	The objective then is to develop policies, $ \{\pi_i: \svec_{i,t}^{jn} \mapsto \vvec_{i,t}, \forall i \}$ that select actions to minimize the expected time to destination while satisfying all the constraints, where $\svec_{i,t}^{jn}$ and $\vvec_{i,t}$ are the joint state and the action of the agent, respectively. Now, the optimization problem can be reformulated as
	\begin{align}
	(\Pb 3): \argmin_{\pi_i} \quad & \E [T_i | \svec_i^{jn}, \pi_j,\forall j\neq i] \notag \\
	s.t.            \quad  &  (\Pb 1.a)-(\Pb 1.c) \notag \\
	& \pvec_{i,t} = \pvec_{i,t-1} + \Delta t \cdot \pi_i(\svec_{i,t-1}^{jn}),  \forall t  \tag{P3.d}
	\end{align}
	where the expectation in the objective function in (P3) is with respect to other agents' unobservable states and policies, and (P3.d) is the agent's kinematics, which satisfy the kinematic constraints in (P1.d) and (P1.e).
	Further, we use the common assumption that each agent would follow the same policy \cite{van2011reciprocal} \cite{chen2016decentralized} \cite{CA_HKretzschmar}, i.e., $\pi = \pi_i$.
	
	\section{Reinforcement Learning Based Approach}	
	In this section, we first introduce reinforcement learning (RL) formulation for the multi-UAV navigation problem. Then, we present the approaches used to tackle the uncertainty in the UAVs' unobservable intents, and the interaction between the UAVs and the cellular network.
	
	\subsection{Reinforcement Learning}	
	 RL is a class of machine learning methods for solving sequential decision making problems with unknown state-transition dynamics \cite{chen2016decentralized}  \cite{everett2020collision}. Typically, a sequential decision making problem can be formulated as an MDP, which is described by the tuple $\langle S,A,P,R,\gamma \rangle $, where $S$ is the state space, $A$ is action space, $P$ is the state-transition model, $R$ is the reward function, and $\gamma$ is a discount factor.

	Since the action space in this work is continuous and the set of permissible velocity vectors depends on the agent's state, we choose to optimize the value function $V_{\pi}(\svec^{jn})$ as in \cite{chen2016decentralized}, instead of optimizing the commonly used action-value function $Q(\svec^{jn}, \vvec)$ (where $\vvec$ denotes the action).
	The state value function of an MDP is the expected return starting from time $t$ following policy $\pi$, i.e.,
	\begin{align}
		V_{\pi}(\svec^{jn}_t) =\sum^T_{t'=t} \gamma^{t'} R_{t'}(\svec^{jn}_{t'}, \pi(\svec^{jn}_{t'})) .
	\end{align}	
	The optimal policy is to maximize the expected return:
	\begin{align}\label{Eq:optimal_policy}
	&\pi^*(\svec_{t}^{jn}) \notag \\
	 &= \argmax_{\vvec_t} R(\svec_{t}^{jn},\vvec_t)
	+ \gamma  \int_{\svec_{t+1}^{jn}} P(\svec_{t+1}^{jn}|\svec_{t}^{jn},\vvec_t)  V^*(\svec_{t+1}^{jn}) \text{d}\svec_{t+1}^{jn},
	\end{align}	
	where $V^* (\svec^{jn}_t) = \sum^T_{t'=t} \gamma^{t'} R_{t'}(\svec^{jn}_{t'}, \pi^*(\svec^{jn}_{t'}))$ is the optimal value function, $R(\svec_{t}^{jn},\vvec_t)$ is the reward received at time $t$, $P(\svec_{t+1}^{jn}|\svec_{t}^{jn},\vvec_t)$ is the transition probability from time $t$ to time $t+1$.

\subsection{Reinforcement Learning Formulation}

 	To estimate the high-dimensional, continuous value function, it is common to approximate it with a deep neural network (DNN) parameterized by weights and biases, $\xibm $.  For notational simplicity, we drop the DNN parameters from the value function notation,  i.e., $\mV(\svec;\xibm ) = \mV(\svec)$. And $\svec$ is the joint state of an agent which is also the input of the DNN, and $\mV(\svec)$ is the output of  the value network given $\svec$.
 
 By detailing each of these elements and relating to (P1.a)-(P1.c) and (P3.d), the following provides an RL formulation for the multi-UAV navigation problem. Each UAV is an independent agent, and in the discussions below,  we use agent instead of UAV.
\subsubsection{State Space}
In multi-agent multi-GBS  cellular networks, the agents are able to observe the following information from the environment: 1) its own information vector $\svec_{i,t}$ (for the $i^{th}$ agent at time step $t$); 2) the observable state of the nearest $J_n < J$ agents $\svec_{i,t}^{jno} = [ \svec_{j,t}^o: j\in \{1,2,...,J_n \} ]$; 3) the location information of the nearest $K_n\leq K$ GBSs, which is assumed to be observed by the agents, and is denoted by $\svec^o_{B} = [\pvec_{B_k}: k \in \{1,...,K_n\} ]  $. 	
All the information observed by the agent constitutes its joint state $\svec_{i,t}^{jn} = [\svec_{i,t}, \svec_{i,t}^{jno}, \svec^o_{B}], \forall t$.

\subsubsection{Action Space}
The action space is a set of permissible velocity vectors. Ideally, the agent can travel in any direction at any time. However, in reality the kinematic constraints in (\ref{Eq:kitc_const})-(\ref{Eq:kitc_const_angl}) restrict the agent's movement and should be taken into account.  Then, based on the agent's current speed, orientation $[v_{s,i,t},\phi_{s,i,t}]$ and the kinematic constraints, permissible actions $\vvec=[v_s, \phi]$ are sampled to built the action space $A_{i,t}$.

\subsubsection{Reward Function}
Similar to  the formulation of the reward function defined in \cite{chen2019crowdrobot},  \cite{chen2016decentralized}, and \cite{everett2020collision}, we define a sparse reward function, which awards the agent for reaching its goal, and penalizes the agent for getting too close or colliding with other agents, and also penalizes for getting close to be disconnected or already being disconnected from the cellular network. The  reward function  consists of three parts: the reward, $R_c$,  that encourages the fast arrival to the destination and penalizes close encounters  with other agents;  the reward, $R_s$, that encourages keeping connectivity with the cellular network, and a constant movement penalty, $R_t$, that encourages the agents to reduce their flight time. For instance, at time step $t$, the reward functions for the $i^{th}$ agent  can be expressed as follows:
\begin{align}
&R_{c_{i,t}}(\svec_{i,t}^{jn}, \vvec_{i,t}) = \notag \\
&\begin{cases}
2,                             & \text{if } \pvec_i = \pvec_i^D, \\
-  (1-  (d_{t_{\min}} - r_i -r_j)/0.2), \\
&\hspace{-0.95in}\text{if } r_i+r_j<d_{t_{\min}} \leq 0.2+r_i + r_j, \\
-1,                         & \text{if } d_{t_{\min}}\leq r_i+r_j,\\
0, & \text{otherwise},
\end{cases} \\
\intertext{and}
&R_{s_{i,t}}(\svec_{i,t}^{jn}, \vvec_{i,t}) = \notag \\
&\begin{cases}
-0.5,  &\text{if }  t \mid n_t \text{ and }  \mT_s \leq \mS_{r_{i,t+1}}< \mT_s + 0.1 \\\
-1,   &\text{if }  t \mid n_t \text{ and }  \mS_{r_{i,t+1}} < \mT_s \\
0, & \text{otherwise},
\end{cases}
\end{align}
where  $d_{t_{\min}}$ is the minimum distance to other agents within the next time step duration.
Therefore, the overall reward function can be expressed as the sum
\begin{align}
&R_{i,t}(\svec_{i,t}^{jn}, \vvec_{i,t}) = R_{c_{i,t}}(\svec_{i,t}^{jn}, \vvec_{i,t}) + R_{s_{i,t}}(\svec_{i,t}^{jn}, \vvec_{i,t}) +R_t.
\end{align}

\subsection{Estimation of the Agents' Unobservable Intents }
 The probabilistic state transition model in (\ref{Eq:optimal_policy}) is determined by the agents' kinematics as defined in (P3.d), other agents' hidden states, and the other agents' choices of action. Since the other agents' hidden intents are unknown, the system's state transition model is unknown as well.
In addition, it is difficult to evaluate the integral, because the other agents' next state has an unknown distribution (that depends on their unobservable intents).  We approximate this integral by assuming that the other agent would be traveling at a filtered velocity for a short duration $\dt$, which is regarded as a one-step lookahead procedure \cite{rvo_JBerg} \cite{van2011reciprocal} \cite{chen2016decentralized}  \cite{hrvo_JSnape}. This propagation step amounts to predicting the other agent's motion with a simple linear model, i.e., $\vhvec_{j,t} = \text{filter}(\vvec_{j,0:t})$. For the $i^{th}$ agent,  other agents' filtered velocities are included in the  vector $\vhvec_{i,t}^{jno} = [\vhvec_{j,t}: j\in \{1,2,...,J_n\}]$.
Then, the estimated next state of the $i^{th}$ agent will be
\begin{align}
\label{Eq:filter_v}
\shvec_{i,t+1,\vvec}^{jn} = [f(\svec_{i,t}, \dt,  \vvec), f(\svec_{i,t}^{jno}, \dt, \vhvec_{j,t}^{jno} ), \svec^o_{B}]
\end{align}
where $f(\cdot)$ is the kinematic model. Then the policy becomes
\begin{align}
\pi^*(\svec_{i,t}^{jn}) &= \argmax_{\vvec} R_{i,t}(\svec_{i,t}^{jn},\vvec)
+ \gamma  V^*(\shvec_{i,t+1,\vvec}^{jn}) .
\end{align}

	\subsection{SINR Prediction}	
	 Model-free RL requires no prior knowledge about the environment. This usually leads to slow learning process and requires a large number of agent-environment interactions, which is typically costly or even risky to obtain \cite{uavtraj_YZeng}. In addition, using only a value network to encode the interactions among the agents and the interactions between the agent and the cellular network is not easy. Actually, each real experience obtained from the agent and cellular network interaction not only can be used to get reward and refine the value network, but also can be used for model learning in order to predict the agent's SINR experienced at certain positions. 	
	More specifically, when flying in the environment, agents interact with the cellular network and obtain the empirical SINR $\widehat{\mS}_r$. Since there is no need to use the exact SINR for connectivity measurement, this work uses the quantized SINR level, $L_w(\widehat{\mS}_r)$,  to check the agent's connectivity.
	With a finite set of measurements $\{\langle \svec_{B}^{jn}, L_w(\svec_{B}^{jn}) \rangle \}$, where $\svec_{B}^{jn} = [\pvec, \svec^o_B]$, a DNN can be trained to predict the SINR level $L_w(\svec_{B}^{jn}) $.
		
	A fully connected DNN with parameters $\xibm _w$ can be used to predict the agent's SINR level, i.e., $\xibm _w$ is trained so that $L_w(\svec_{B}^{jn})  \approx \mL_w(\svec_{B}^{jn}; \xibm _w)$. The data measurement $\langle \svec_{B}^{jn}, L_w(\svec_{B}^{jn})  \rangle$ only arrives incrementally as the agent flies to new locations and can be saved in a database (e.g., replay memory), and a minibatch is sampled at random from the database to update the network parameter $\xibm _w$.
	Note that the prediction of SINR levels might be highly inaccurate initially, but can be continuously improved as more real experience is accumulated.
	

	\section{Decentralized Deep Reinforcement Learning Algorithm}
	In this section, we present the proposed  decentralized deep reinforcement learning algorithm as a solution to multi-UAV navigation with collision avoidance and wireless connectivity constraints, including the SINR-prediction neural network. The proposed algorithm is presented in Algorithm \ref{Algm:main_algm}, and is referred to as RLTCW-SP (RL for Trajectory optimization with Collision avoidance and Wireless connectivity constraint and with SINR Prediction).
	
	\begin{algorithm}
		\caption{RLTCW-SP Algorithm}
		\label{Algm:main_algm}
		\LinesNumbered
		\KwIn{State-value pairs $D$}
		Initialize state-value pairs  $D$\\
		Initialize location-SINR pairs $D_w$\\
		Initialize value network $\xibm$ with $D$\\
		Initialize SINR-prediction network $\xibm_w$	\\		
		\For{episode = 0: total episode}{
			\For{n random training cases}{
			Initialize  $\svec_{i,0} \forall i$ \\
			\While{not all reached destinations}{
				\For{each agent $i$}{
					\If{not reached destination}{
						$\svec_{i,t}^{jn} \leftarrow \text{observeEnvironment}()$ \\
						$A_{i,t} \leftarrow \text{sampleActionSpace}()$ \\
						$c \leftarrow \text{randomSample(Uniform (0,1))}$\\
						\eIf{ $c\leq \epsilon$ }{
							$\vvec_{i,t} \leftarrow \text{randomSample} (A_{i,t})$ }{
							$\vhvec_{i,t}^{jno} \leftarrow \text{filter}(\vvec_{0:t-1}^{jn})$\\
							$\shvec_{i,t+1}^{jno} \leftarrow \text{propagate}(\svec_{i,t}^{jno},
							\vhvec_{i,t}^{jno})$ \\
							\For{every $\avec$ in $A_{i,t}$}{
								$\shvec_{i,t+1} \leftarrow \text{propagate}(\svec_{i,t},
								\avec)$ \\
								$\hat{L}_{w_{i,t+1}} = \mL_w(\shvec_{B_{i,t+1}}^{jn})$\\
								$R_{i,t} \leftarrow   \text{getReward}(\shvec_{i,t+1}^{jn}, \hat{L}_{w_{i,t+1}})$\\
								$V_{p} = R_{i,t} + \gamma \mV (\shvec_{i,t+1}^{jn})$
							} 
							$\vvec_{i,t}  \leftarrow \argmax_{\avec\in A_{i,t}} V_{p}$
						} 
						$R_{i,t}, \svec_{i,t+1}, \mS_{r_{i,t+1}} \leftarrow \text{executeAction}(\vvec_{i,t} )$						
					}}}	
				\For{each agent $i$}{		
					$V_{i,0:T_i} \leftarrow \text{updateValue}(\svec^{jn}_{i,0:T_i},R_{i,0:T_i},\xibm) $		\\
					$L_{w_{i,0:T_i}}  \leftarrow \text{getSINRlevel}(\mS_{r_{i,0:T_i}})$ \\
					Update state-value pairs $D$ with $\langle\svec_{i,0:T_i}^{jn}, V_{i,0:T_i} \rangle$\\		
					Update location-SINR pairs $D_w$ with $\langle\svec_{B_{i,0:T_i}}^{jn},L_{w_{i,0:T_i}}  \rangle$
				}
			}
			Sample random minibatch from $D$, and update value network $\xibm$ by gradient descent.\\
			Sample random minibatch from $D_w$, and update SINR-prediction network $\xibm_w$ by gradient descent.
		}
		\Return $\xibm, \xibm_w$
	\end{algorithm}
	
	\subsection{Parametrization}
	Since the optimal policy should be invariant to any coordinate plane, we follow the agent-centric parameterization as in \cite{chen2019crowdrobot}, \cite{chen2016decentralized} and \cite{everett2020collision}, where the agent is located at the origin and the $x$-axis is pointing toward the agent's destination. The states of the $i^{th}$ agent after transformation is
	\begin{align}
	&\stvec_i =  [d_{g_i}, v_{\max_i}, \tilde{v}_{x_i}, \tilde{v}_{y_i}, r_i, \tilde{\phi}_i] \\
	& \stvec^{jno}_i =   [[\tilde{p}_{x_j}, \tilde{p}_{y_j}, H_V, \tilde{v}_{x_j}, \tilde{v}_{y_j}, r_j, d_{j}]: j \in \{ 1,2,...,J_n  \} ]
	\end{align}
	where $d_g$ is the agent's distance to the goal, $d_j$ is the agent's distance to the $j^{th}$ agent, and $\tilde{p}$ denotes $p$ in the new coordinate.
	
	In addition, SINR experienced at an agent depends on the distance and the relative angular direction from the agent to the GBSs, while it does not depend on the positions in global coordinates.  To remove this redundant dependence, the location information vector of all GBSs can be parameterized as
	\begin{align}
	& \ptvec_{B_k}   = [d_{B_k}, \phi_{B_k}, \theta_{B_k}] \\
	&\stvec_{B_i}  = [\ptvec_{B_k}: k \in \{1,...,K_n\}   ]
	\end{align}
	where $d_{B_k} = || \pvec_{B_k} - \pvec_i ||$ is the distance from the agent to the $k^{th}$ BS, $\phi_{B_k}$ and $\theta_{B_k}$ are the horizontal and vertical angles of the $k^{th}$ BS with respect to the agent.
	
	Therefore, the joint state of the $i^{th}$ agent after transformation is
	\begin{align}
	\stvec^{jn}_i = [\stvec_i, \stvec^{jno}_i, \stvec_{B_i}].
	\end{align}
	And the input of  the SINR-prediction network becomes
	\begin{align}
	\stvec^{jn}_{B_i} =  [[d_{B_k}, \phi_{B_k}, \theta_{B_k}]: k \in \{1,...,K_n\}   ].
	\end{align}

	\subsection{Initialization}
	The value network $\xibm$ can be  first initialized with imitation learning using a set of experiences to accelerate the convergence. More specifically, in this work, we use optimal reciprocal collision avoidance (ORCA) \cite{van2011reciprocal} to generate  a number of  trajectories that contain a large set of state-value pairs $\{\langle\svec^{jn} , V  \rangle \}^{N_1}$, where $V = \gamma^{t_g} $ and $t_g$ is the time to reach the destination. The experiences are saved in memory $D$ (line 1 in Algorithm \ref{Algm:main_algm}).
	Then, the value network is initialized by supervised training on $D$ (line 3). The value network is trained by back-propagation to minimize a quadratic regression error
	\begin{align}
	\xibm = \argmin_{\xibm' } \sum_{k=1}^{N_1} \left(V_k - \mV(\svec^{jn}_k ; \xibm' )  \right).
	\end{align}
	
	If a set of location-SINR experiences can be downloaded from the cloud, we can save the downloaded dataset in memory $D_w$ (line 2), $\{ \langle \svec^{jn}_{B}  ,L_w\rangle  \}^{N_2}$, where $L_w$ is the scaled SINR level that the agent experienced.  Then, the SINR-prediction network can be initialized with  $\xibm_w = \argmin_{\xibm'} \sum_{k=1}^{N_2} \left(L_{w_k} - \mL(\svec^{jn}_{B}; \xibm') \right)$, which is trained by back-propagation (line 4). If no dataset is available, $D_w$ is initialized with an empty list, and the SINR-prediction network is initialized with random network parameters.

	\subsection{Refining Process}
	After initialization, a refining process is performed using RL.
	Particularly, a set of random training cases is generated in each episode (line 6). In each training case, each agent navigates around others to arrive its destination, while interacting with the cellular network (line 10- line 25). It is worth noting that the agents navigate simultaneously and with no communication among each other. At each time step $t$, each agent first observes the environment, obtains the observable states of other nearby agents and the location information of the GBSs, and then obtains its joint state $\svec^{jn}_{t}$ (line11). Then, based on its current velocity and kinematic constraints, each agent builds an action space $A_{t}$ (line 12).
	Using an $\epsilon$-greedy policy, each agent selects a random action with probability $\epsilon$ from $A_{t}$ (line 15), or follows the value network greedily otherwise (lines 17-24). When following the value network to choose actions, each agent performs the following: 1) estimate other nearby agents' motion by filtering their velocities, and estimate their observable states $\shvec^{jno}_{t+1}$ following equation (\ref{Eq:filter_v})  (lines 17-18); 2) predict its next SINR level $\mL_{w_{t+1}}$ using the SINR-prediction network $\xibm_w$; 3) choose the best action in $A_t$ which has the maximum $V_p$.

	When all agents have arrived their destinations in each training case,  trajectories $\svec_{i, 0:T_1}\forall i$ are then processed to generate a set of state-value pairs $\langle\svec_{i,0:T_i}^{jn}, V_{i,0:T_i} \rangle$, where
	\begin{align}
	V_{i,t} =
	\begin{cases}
	R_{i,t} + \gamma \mV (\svec_{i,1:t+1}^{jn}) &\text{if } t <T_i, \\
	R_{i,t} &\text{if } t =T_i,
	\end{cases} \notag
	\end{align}	
	and a set of location-SINR pairs $\langle[\pvec_{i,0:T_i}, \svec^o_B],L_{w_{i,0:T_i}}  \rangle$. The new pairs are used to update $D$ and $D_w$.

	\subsection{Training}
	We first use optimal reciprocal collision avoidance (ORCA) \cite{van2011reciprocal} to generate  a number of  trajectories that contain a large set of state-value pairs $\{\langle\svec^{jn} , V  \rangle \}^{N_1}$, where $V = \gamma^{t_g} $ and $t_g$ is the time to reach the destination. The experiences, as input of Algorithm \ref{Algm:main_algm}, are saved in memory $D$, which will be refined during training.
	To train the value network and SINR-prediction network, a set of training points is randomly sampled from the experience set, which contains state-value pairs for $\xibm$ or location-SINR pairs for $\xibm_w$ from many different trajectories. Then, the networks are finally updated by stochastic gradient descent (back-propagation) on the sampled subsets of experience.
	
	\subsection{Real-Time Navigation}
	With the trained value network and SINR-prediction network, agent can execute real-time navigation. This process is provided in Algorithm \ref{Algm:execution}.
	
	\begin{algorithm}
		\caption{Real-Time Navigation}
		\label{Algm:execution}
		\LinesNumbered
		\KwIn{$\xibm, \xibm_w$}
		Initialize $\svec_{0}$ \\
		\While{not reached destination}{
					$\svec_{t}^{jn} \leftarrow \text{observeEnvironment}()$ \\
					$A_{t} \leftarrow \text{sampleActionSpace}()$ \\					
					$\vhvec_{t}^{jn} \leftarrow \text{filter}(\vvec_{0:t-1}^{jn})$\\
					$\shvec_{t+1}^{jno} \leftarrow \text{propagate}(\svec_{t}^{jno},
					\vhvec_t^{jn})$ \\
					\For{every $\avec$ in $A_{t}$}{
						$\shvec_{t+1} \leftarrow \text{propagate}(\svec_{t},
						\avec)$ \\
						$\hat{L}_{w_{t+1}} = \mL_w([\phvec_{t+1}, \svec_B])$\\
						$R_{t} \leftarrow   \text{getReward}(\shvec_{t+1}, \shvec_{t+1}^{jno}, \hat{L}_{w_{t+1}})$\\
						$V_{p} = R_{t} + \gamma \mV (\shvec_{t+1}^{jn})$
					} 
					$\vvec_{t}  \leftarrow \argmax_{\avec\in A_{t}} V_{p}$ \\
					$\svec_{t+1}\leftarrow \text{executeAction}(\vvec_{t} )$											
		}
		\Return $\vvec_{0:T-1}$, $\svec_{0:T}$
	\end{algorithm}

		\begin{figure}		
		\centering
		\begin{minipage}{0.45\textwidth}
			\centering
			\includegraphics[width=1\textwidth]{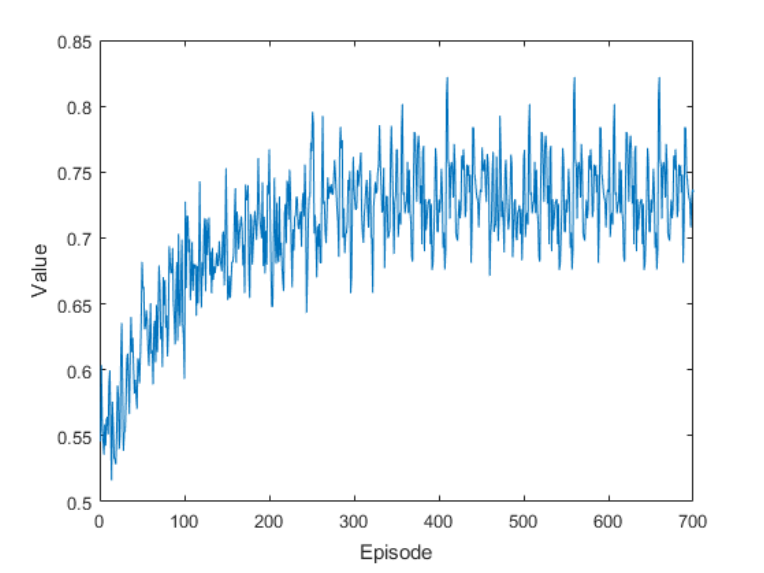}
			\subcaption{\scriptsize Value of the value netowrk. }
		\end{minipage}	
		\begin{minipage}{0.45\textwidth}
			\centering
			\includegraphics[width=1\textwidth]{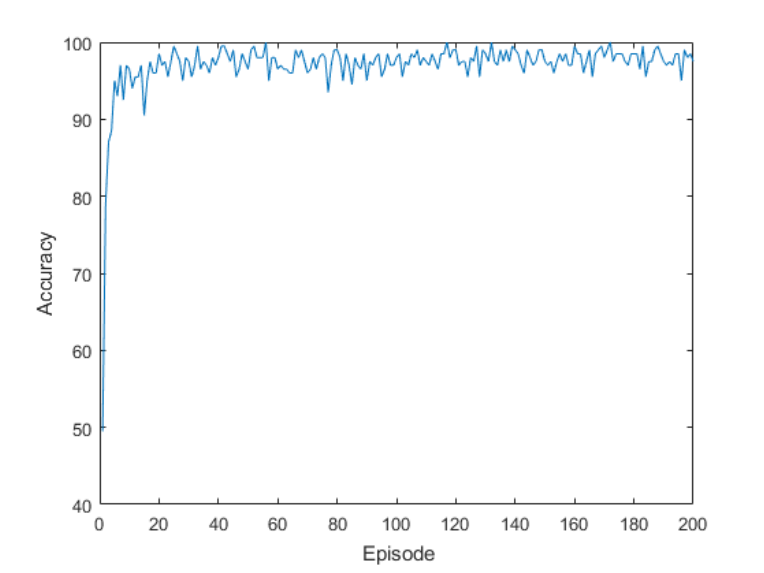}
			\subcaption{\scriptsize Accuracy of the SINR-prediction network.}
		\end{minipage}			
		\caption{\small Value of the value network and accuracy of the SINR-prediction network as functions of the number of episodes.  \normalsize}
		\label{Fig:convergence}
	\end{figure}

	\section{Numerical Results}
	In this section, we present the numerical results to evaluate the performance of the proposed algorithms.
	In the illustrations of environment and trajectories in this section, the GBSs are marked by blue triangles, and the yellow areas indicate the communication coverage zones where the agents are able to connect with the cellular network (i.e., $\mS_{r} \geq \mT_{s} $). Agents' trajectories are displayed as a list of dots in different colors, and the destinations are marked with crosses.
	In each flight trajectory, there are four possible outcomes for the agent/UAV: 1) success, if the agent arrives its destination successfully; 2) collision, if it collides with others; 3) disconnection, if the continuous disconnected time is larger than the threshold $\mT_t$; 4) stuck, if the agent freezes and stops moving and consequently does not reach the destination.  In addition, we also compute the additional average time (referred to also as average more time) needed to reach the destination, when compared with the lower bound (attained when the UAV goes straight towards the destination at the maximum speed). Therefore, we use success rate (SR), collision rate (CR), disconnection rate (DR) and average more time (AMT) to show the performance of the algorithms.

	\subsection{Environment Setting and the Networks}
	Since the agents fly at the same altitude, the area of interest becomes two-dimensional.
	In the simulations, we consider an area  with 12 GBSs deployed. The GBSs transmit with power $P_B=1$ dBW, have a height of $H_B=32$ m, and the antenna patterns are set with $\theta^{tilt}=10^{\circ}$ and $\theta^{3dB} = 15^{\circ}$. The UAVs are assumed to fly at a fixed altitude of $H_V=50$ m. The noise power is $\mathcal{N}_s = 10^{-6}$, and the SINR threshold is $\mT_{s} = -3$ dB.  Each UAV, as an independent agent, is able to observe the nearest 8 GBSs' locations and at most 4 other agents' observable states.
	
	We construct the value network via a three-layered DNN of size (64,32,16). The exploration parameter $\epsilon$ linearly decays from $0.5$ to $0.1$. The replay memory capacity is 30000 for the 2-agent scenario and 100000 for scenarios with more than two agents. The SINR-prediction network is constructed via a three-layered DNN of size (32,16,8). A standardization layer is utilized after the input layer of both networks.  ReLU activation function is used for the input layer and two hidden layer for both networks. 
	Both networks use Adam optimizer, and have learning rate 0.01, batch size 200, and a regularization parameter 0.0001.

	\begin{figure*}
		\centering
		\begin{minipage}{0.23\textwidth}
			\centering
			\includegraphics[width=1\textwidth]{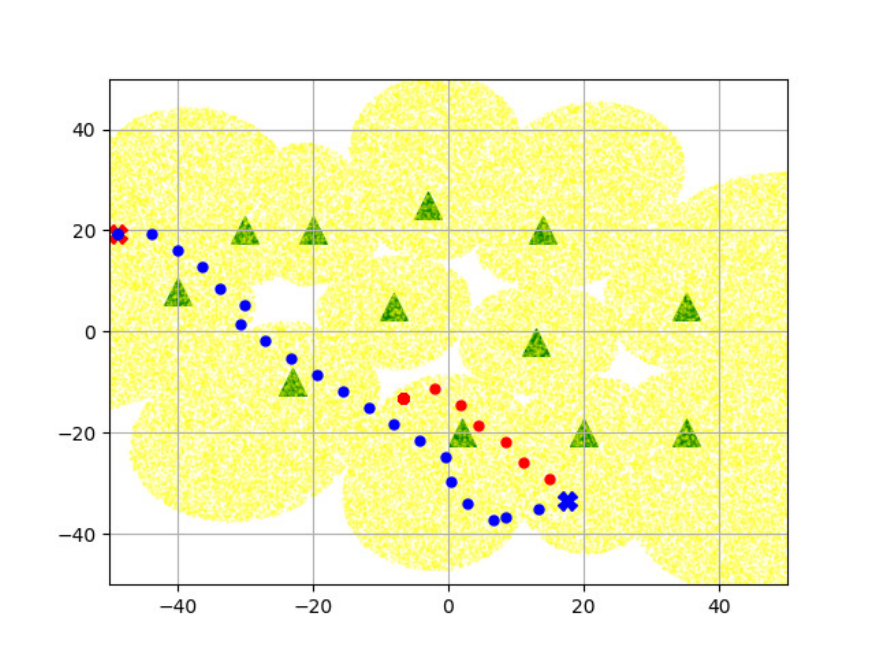}
			\subcaption{\scriptsize Episode 0. }
		\end{minipage}
		\begin{minipage}{0.23\textwidth}
			\centering
			\includegraphics[width=1\textwidth]{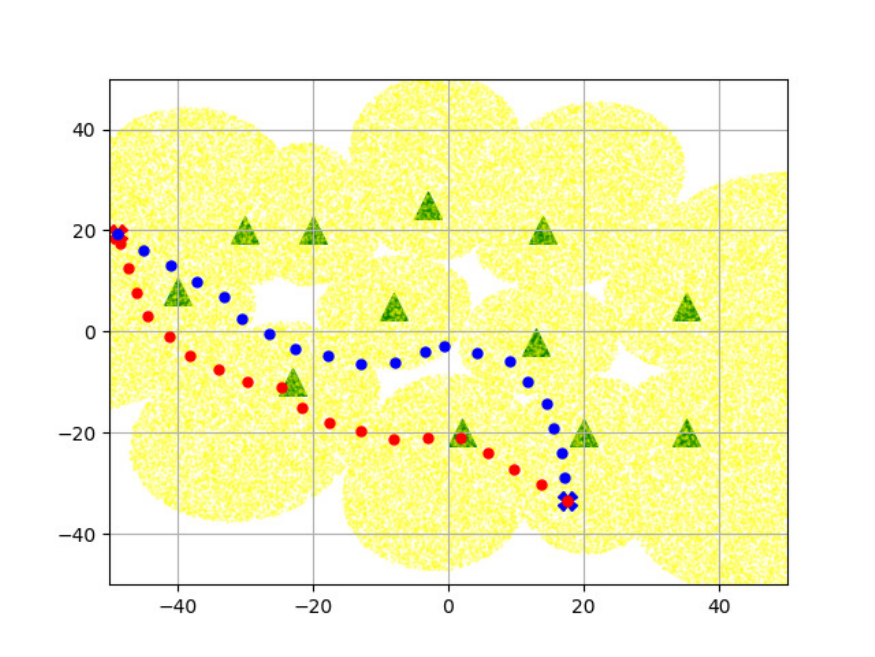}
			\subcaption{\scriptsize Episode 100. }
		\end{minipage}
		\begin{minipage}{0.23\textwidth}
			\centering
			\includegraphics[width=1\textwidth]{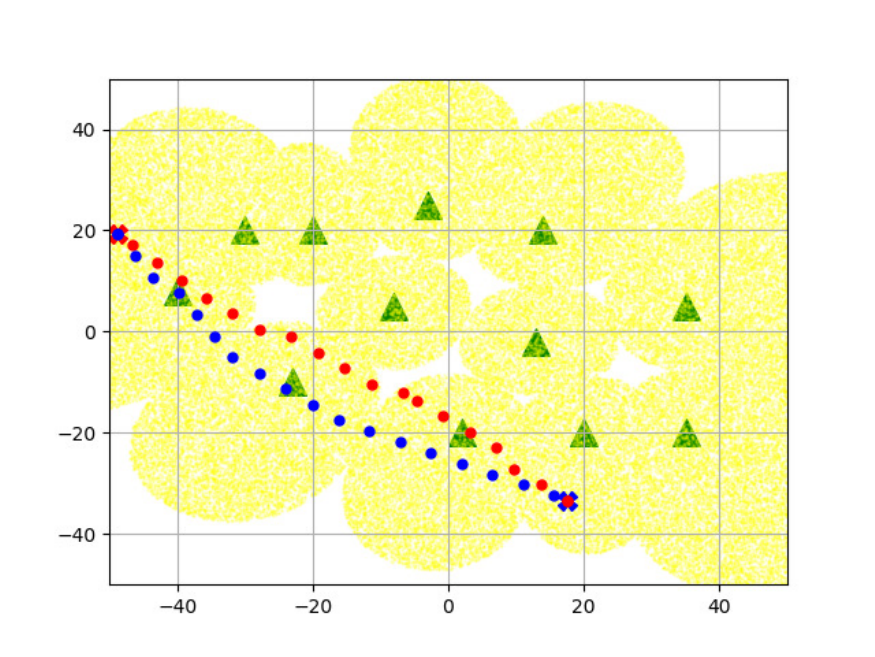}
			\subcaption{\scriptsize Episode 200. }
		\end{minipage}
		\begin{minipage}{0.23\textwidth}
			\centering
			\includegraphics[width=1\textwidth]{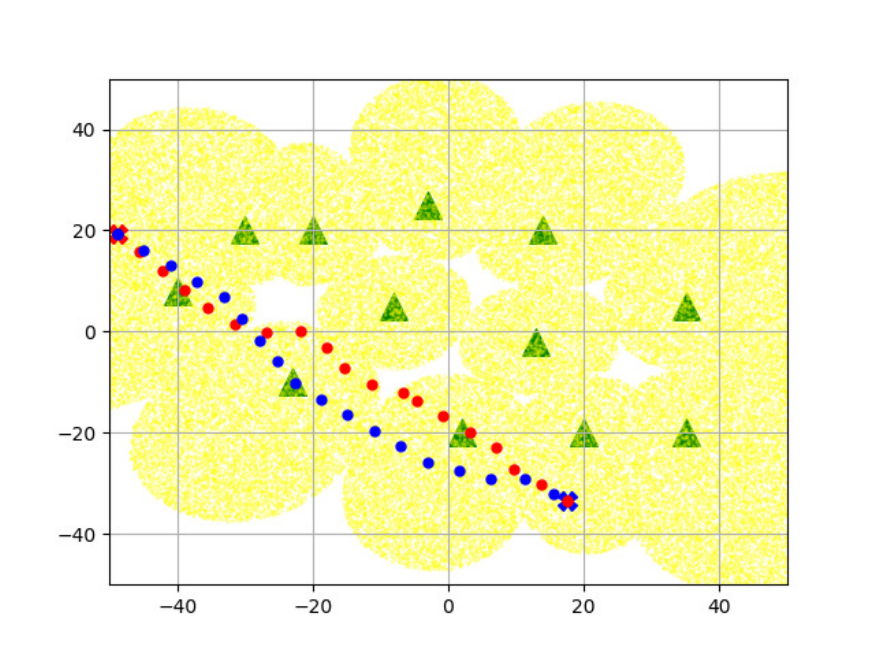}
			\subcaption{\scriptsize Episode 300. }
		\end{minipage}
		\begin{minipage}{0.23\textwidth}
			\centering
			\includegraphics[width=1\textwidth]{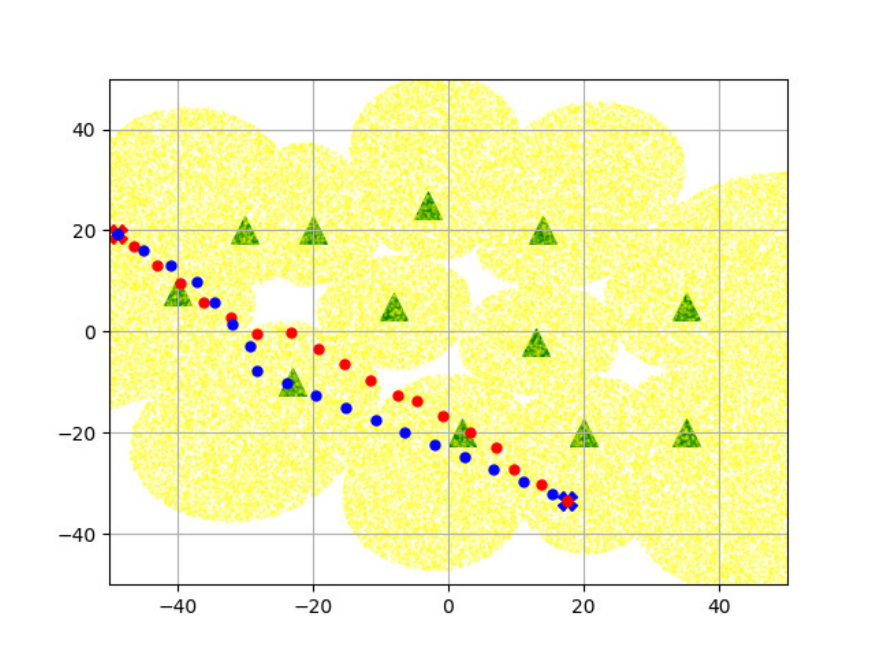}
			\subcaption{\scriptsize Episode 400. }
		\end{minipage}
		\begin{minipage}{0.23\textwidth}
			\centering
			\includegraphics[width=1\textwidth]{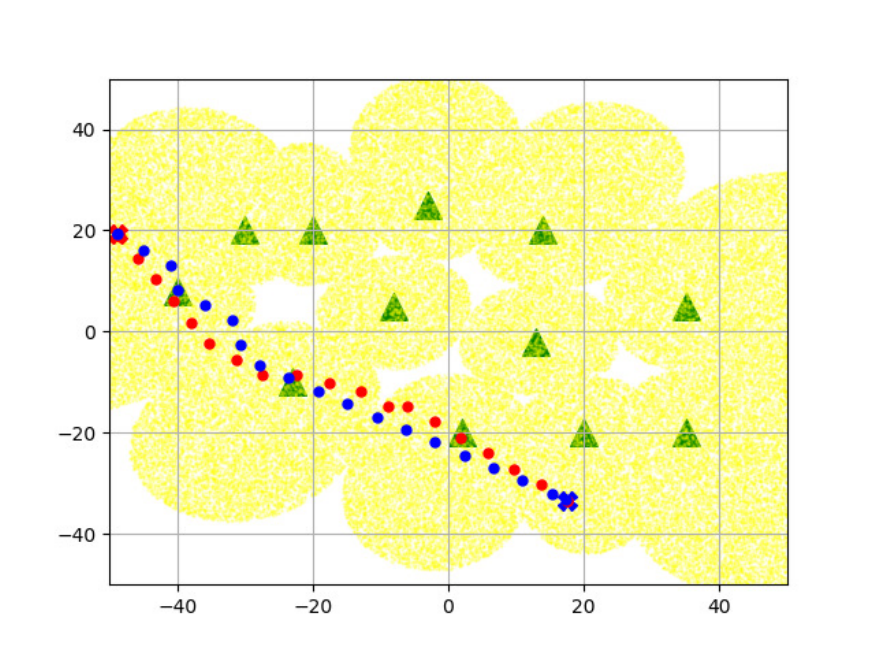}
			\subcaption{\scriptsize Episode 500. }
		\end{minipage}
		\begin{minipage}{0.23\textwidth}
			\centering
			\includegraphics[width=1\textwidth]{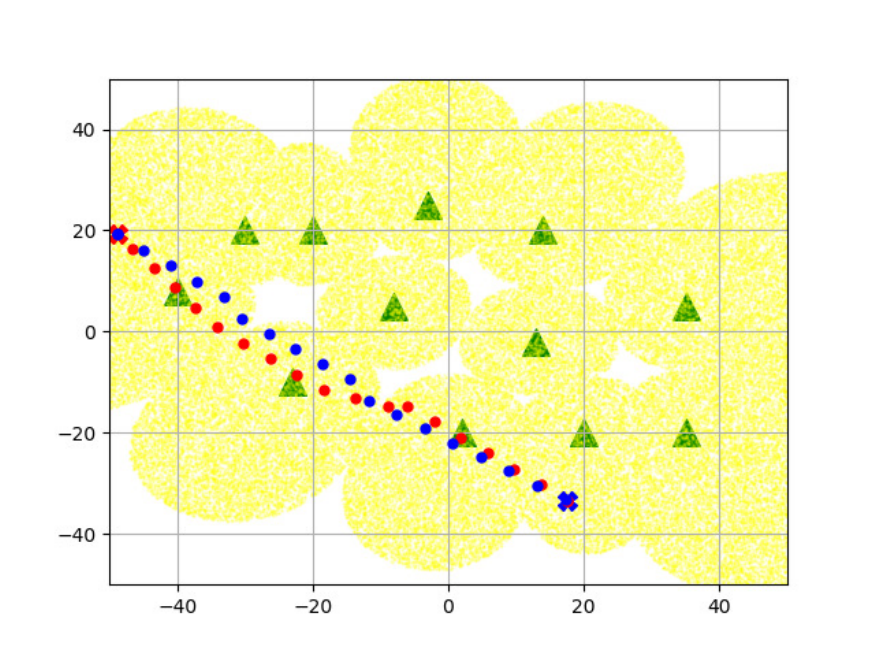}
			\subcaption{\scriptsize Episode 600. }
		\end{minipage}
		\begin{minipage}{0.23\textwidth}
			\centering
			\includegraphics[width=1\textwidth]{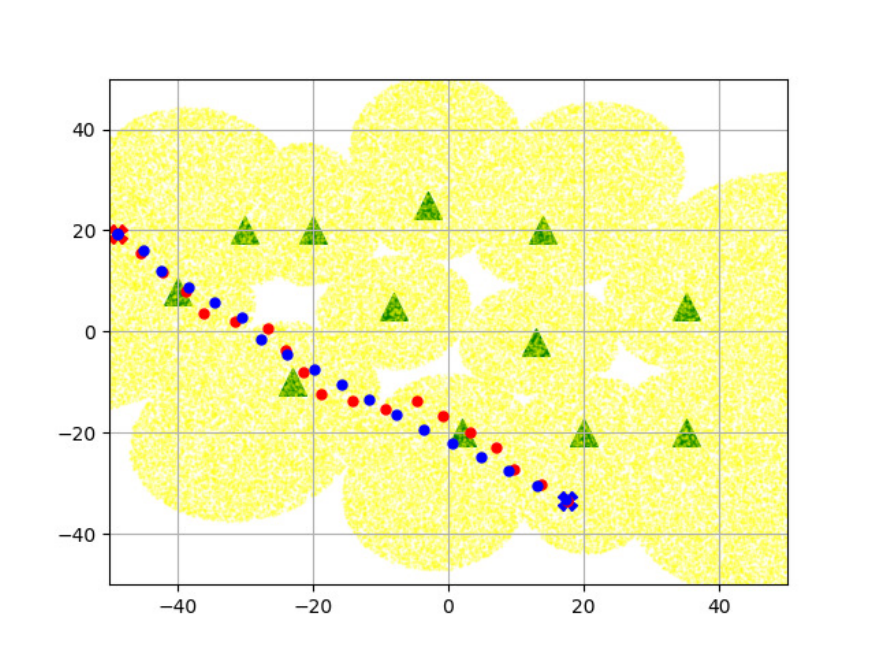}
			\subcaption{\scriptsize Episode 700. }
		\end{minipage}
		\caption{\small Trajectory examples at different episodes during training.  \normalsize}
		\label{Fig:traj_optimization}
	\end{figure*}

	\subsection{Convergence in Training}
	Fig. \ref{Fig:convergence} shows the value of the value network and accuracy of the SINR-prediction network as functions of the number of episodes during training for a  2-agent scenario.
	Fig. \ref{Fig:convergence}(a) shows that the value converges after around 200 episodes.
	From Fig. \ref{Fig:convergence}(b), we can see that the accuracy converges after around 20 episodes, since in each episode 50 random trajectories are generated for each agent, during which more than 15000 location-SINR pairs are collected and used to train the SINR-prediction network.

	The trajectory optimization process for two UAVs is displayed in Fig. \ref{Fig:traj_optimization}. At episode 0, the SINR-prediction network is initialized with random weights and bias, and is not able to predict the accurate SINR level. Besides, the policy has not been refined by RL. As a result, the two agents are easily getting  disconnected or stuck.  After 100 episodes of training, the SINR-prediction network is well-trained and able to predict the SINR levels with 97\% accuracy. Also, the value network is trained with refined state-value pairs. Thus, the agents can reach their destinations, but with long trajectories to avoid collisions and disconnection. As the training proceeds,  the policy improves, leading to shorter expected trajectories.  Table \ref{Table:traj_optimiztion} presents the AMT (for the successful trajectories) in different episodes, and we clearly observe the declining AMT values.
	
	Separately,  we also compute the  AMT  in two different scenarios for comparison: 1) the connectivity constraint is not considered for the two-agent trajectory design (CADRL \cite{chen2016decentralized}); and 2) the collision avoidance constraint is not considered if there is only one agent. The AMT in these two scenarios are 0.578s, and 0.354s, respectively.

	\begin{table*}
		\centering
		\caption{Average more time need to reach destination at different episodes.}
		\label{Table:traj_optimiztion}
		\begin{tabular}{|c|c|c|c|c|c|c|c|c|}
			\hline
			Number of Episodes      & 0 & 100   & 200   & 300    & 400    & 500   & 600   & 700   \\ \hline
			Average More Time (s) & 2.45  & 1.723 & 0.972 & 0.8707 & 0.8007 & 0.772 & 0.766 & 0.712 \\ \hline
		\end{tabular}
	\end{table*}

    \subsection{Testing of Navigation in Different Environments}
        \begin{figure*}
    	\centering
    	\begin{minipage}{0.32\textwidth}
    		\centering
    		\includegraphics[width=1\textwidth]{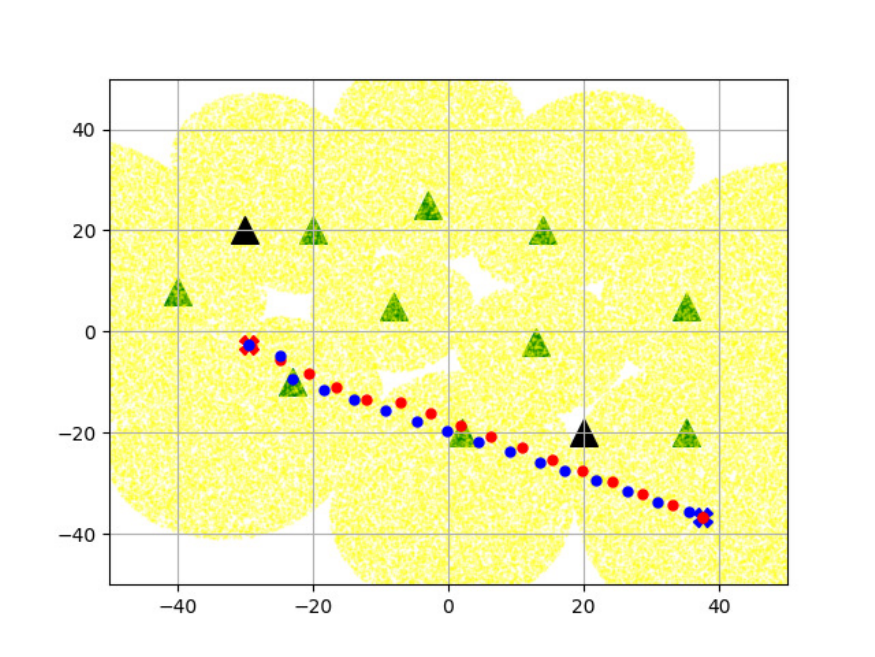}
    		\subcaption{\scriptsize DE1.   }
    	\end{minipage}
    	\begin{minipage}{0.32\textwidth}
    		\centering
    		\includegraphics[width=1\textwidth]{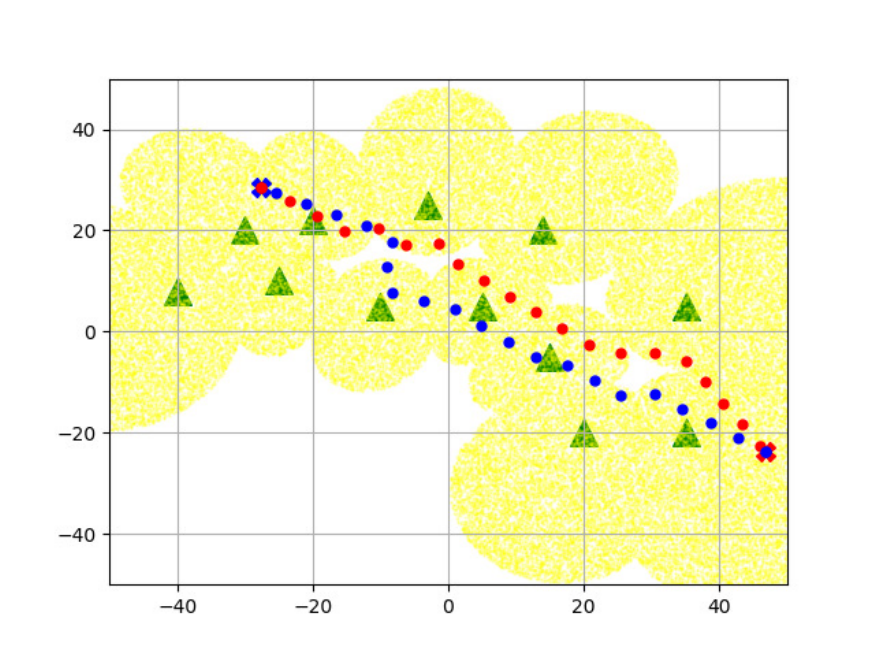}
    		\subcaption{\scriptsize DE2.}
    	\end{minipage}
	    \begin{minipage}{0.32\textwidth}
	    	\centering
	    	\includegraphics[width=1\textwidth]{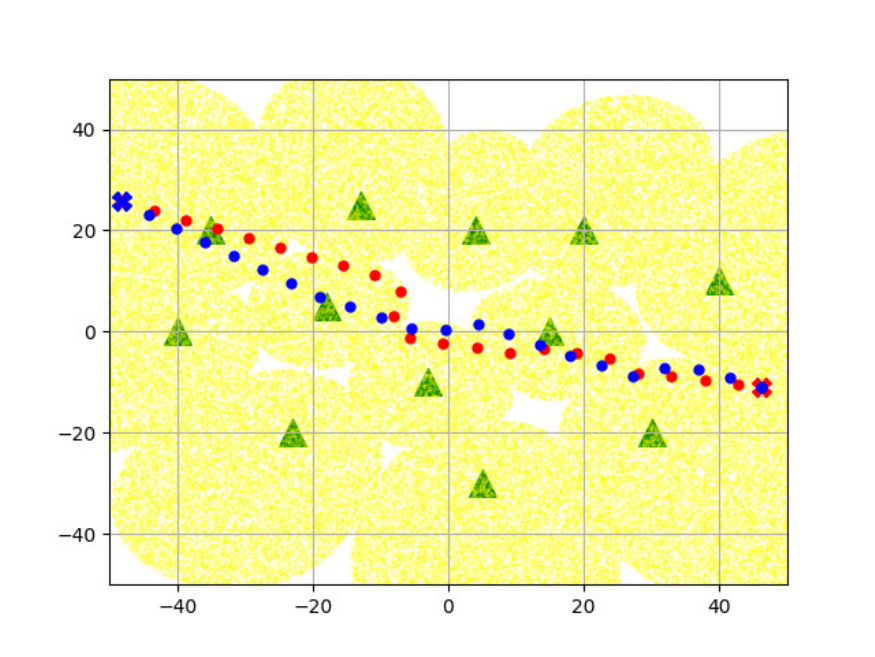}
	    	\subcaption{\scriptsize DE3.}
	    \end{minipage}
    \begin{minipage}{0.32\textwidth}
    	\centering
    	\includegraphics[width=1\textwidth]{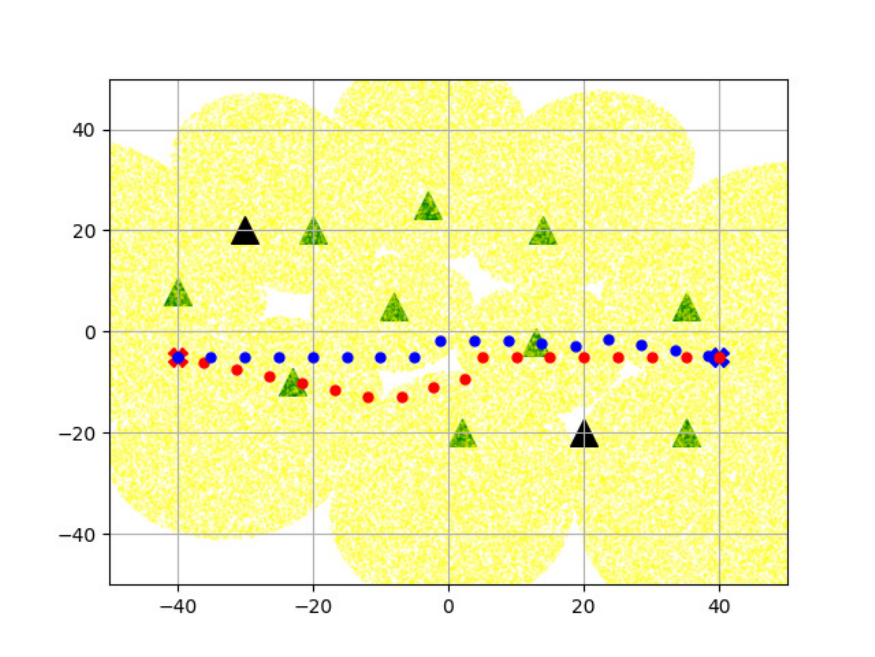}
    	\subcaption{\scriptsize DE1.   }
    \end{minipage}
    \begin{minipage}{0.32\textwidth}
    	\centering
    	\includegraphics[width=1\textwidth]{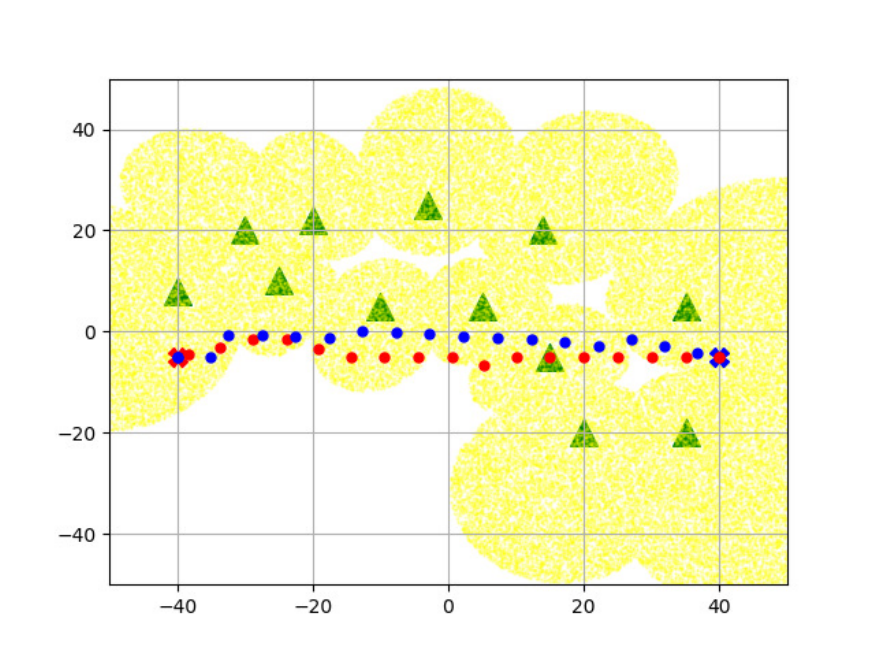}
    	\subcaption{\scriptsize DE2.}
    \end{minipage}
    \begin{minipage}{0.32\textwidth}
    	\centering
    	\includegraphics[width=1\textwidth]{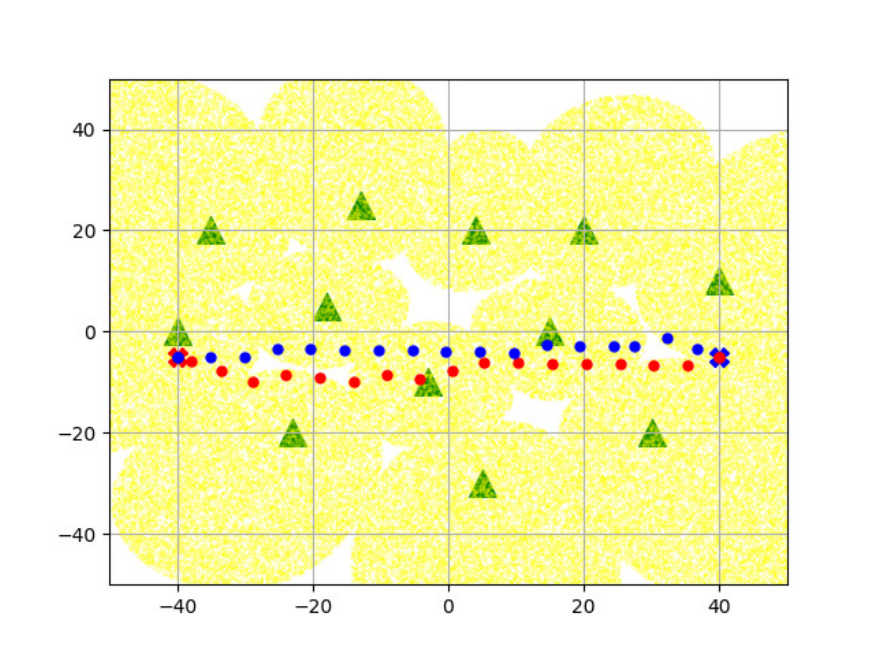}
    	\subcaption{\scriptsize DE3.}
    \end{minipage}
    	\caption{\small Illustrations of different environments used in navigation testing, and trajectory examples when using proposed RLTCW-SP algorithm. In (a) and (d), the two BSs in black are not operational for the UAVs.
    		\label{Fig:diffenv_comparison}  \normalsize}
    \end{figure*}
    In the proposed RLTCW-SP algorithm, an SINR-prediction network is trained to predict the SINR level. In an ideal scenario, the antenna pattern information of GBSs may be available to the agents, and then the agents are able to predict the SINR with that information. In this subsection, we compare the performances of the following three algorithms: 1) the agents are able to get the antenna pattern information of the GBSs, and then predict the SINR directly (referred to as RLTCW-AW); 2) the proposed RLTCW-SP algorithm, which uses the location-SINR memory, and trains an SINR-prediction network to predict the SINR level; 3) the agents do not predict the SINR and only use the value network to encode the interaction with the cellular network (referred to as RLTCW). The navigation test is done in three types of environments: 1) the same environment as in the training; 2) the same environment but two BSs are not operational for the UAV (due to congestion, malfunction,  resource allocation to ground users, or GBS activation schedule), an illustration of which is presented in Fig. \ref{Fig:diffenv_comparison}(a); 3) different environment with different GBS deployment, illustrations of which are presented in Figs. \ref{Fig:diffenv_comparison}(b) and \ref{Fig:diffenv_comparison}(c). Fig. \ref{Fig:diffenv_comparison} also presents examples of trajectories that the agents perform using the proposed RLTCW-SP algorithm in a challenging scenario in which the destination of one UAV is the starting point of the other UAV. Environments displayed in Figs. \ref{Fig:diffenv_comparison} (a) (b) and (c) are referred as DE1, DE2 and DE3 (using DE as the abbreviation for different environment).

    The performance of the three algorithms in different environments are presented in Table \ref{Table:diffenv_comparison}. As expected, the RLTCW-AW with the perfect knowledge of antenna patterns has the best performance, and the RLTCW-SP algorithm has slightly lower performance which is due to the potential inaccuracies  in the SINR prediction, while the performance of RLTCW is substantially lower compared to the other two, due to very high DR (disconnection rate). In addition, the SR (success rate) performance of the proposed RLTCW-SP algorithm decreases only slightly in different environments, and how large the decrease is depends on which environment is used in testing. When there are large and wide out-of-coverage zones in the environment (as shown in Fig. \ref{Fig:diffenv_comparison}(b)), the SR performance decreases more. The reason is that the wide out-of-coverage zones are more likely to make the agent get stuck at the edge and not be able to decide which direction to go.


    	\begin{table*}[t]    		
    		\centering
    		\caption{Performance of different algorithms in different environments in terms of success rate (SR), collision rate (CR), and disconnection rate (DR) \textbf{(all rates are in \%)}. }
    		\label{Table:diffenv_comparison}
    		\begin{tabular}{|c|c|c|c|c|c|c|c|c|c|c|c|c|}
    			\hline
    			\multirow{2}{*}{} & \multicolumn{3}{c|}{Same Environment} & \multicolumn{3}{c|}{DE1} & \multicolumn{3}{c|}{DE2}& \multicolumn{3}{c|}{DE3} \\ \cline{2-13}
    			&SR  &CR  &DR &SR  &CR  &DR &SR  &CR  &DR &SR  &CR  &DR    \\ \hline
    			RLTCW-AW   &94.8 &2.8 &0 &95 &5 &0 &87.6 &6.2 &0.6 &92.2 &6.8 &0    \\ \hline
    			RLTCW-SP &94.02 &4 &0.06 &93.4 &5.8 &0.07 &86.1 &7.4 &  0.7   &91.8&6.2&0.1    \\ \hline
    			RLTCW   & 70.4        &  3.4      & 23.4       &  77.95          & 3.3           & 18.2           &  59.3       &  4.8        & 28.4  &70.6&4.2& 22.9   \\ \hline
    		\end{tabular}
    	\end{table*}
    
        \begin{figure*}
        	  	\centering    	
    	\begin{minipage}{0.32\textwidth}
    		\centering
    		\includegraphics[width=1\textwidth]{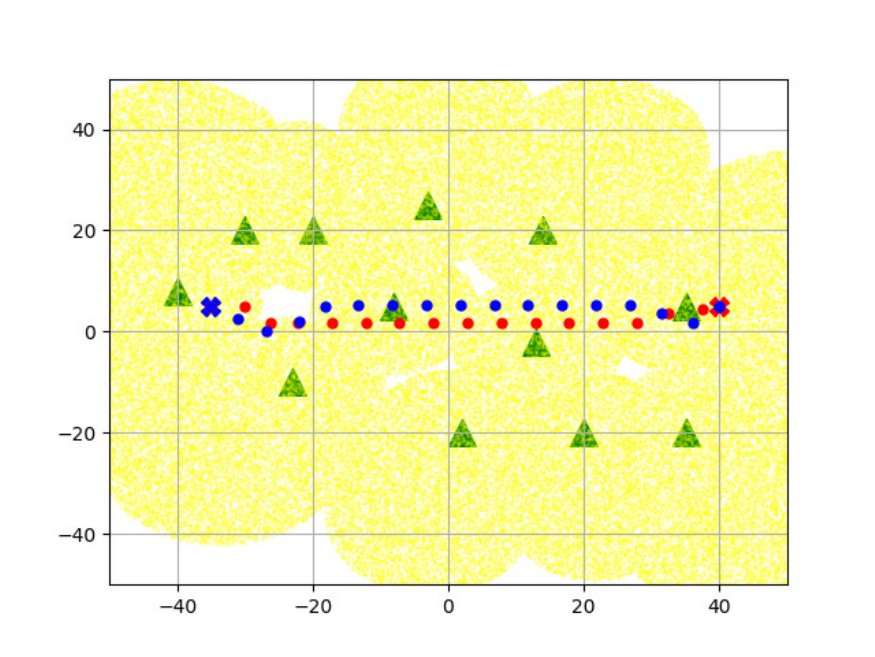}
    		\subcaption{\scriptsize  }  	
    		\label{Fig:diff_setting_original-1}	
    	\end{minipage}
    	\begin{minipage}{0.32\textwidth}
    		\centering
    		\includegraphics[width=1\textwidth]{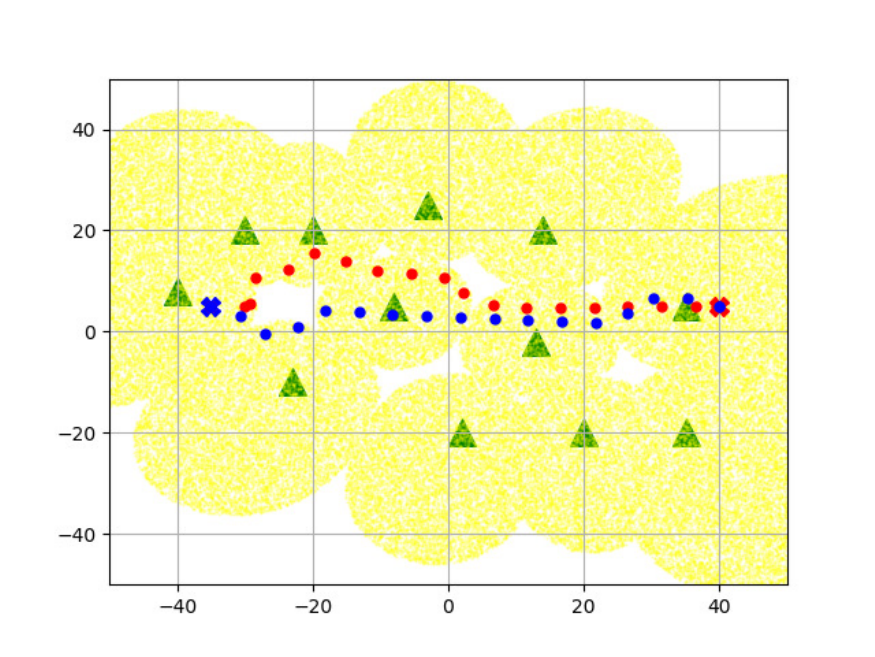}
    		\subcaption{\scriptsize  }
    		\label{Fig:diff_setting_H100-1}	
    	\end{minipage}
    	\begin{minipage}{0.32\textwidth}
    		\centering
    		\includegraphics[width=1\textwidth]{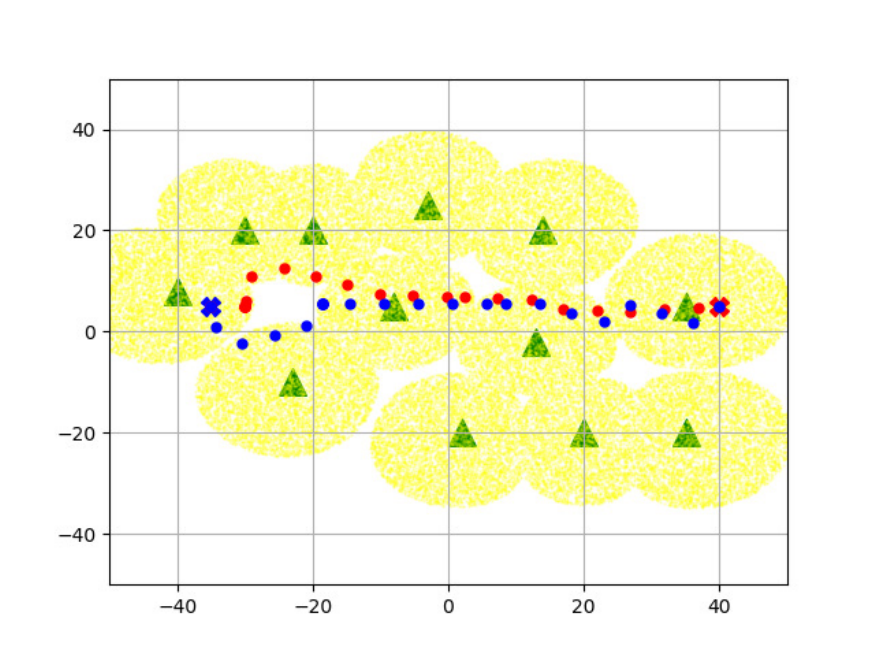}
    		\subcaption{\scriptsize  }
    		\label{Fig:diff_setting_theta15-1}	
    	\end{minipage}
    	\begin{minipage}{0.32\textwidth}
    		\centering
    		\includegraphics[width=1\textwidth]{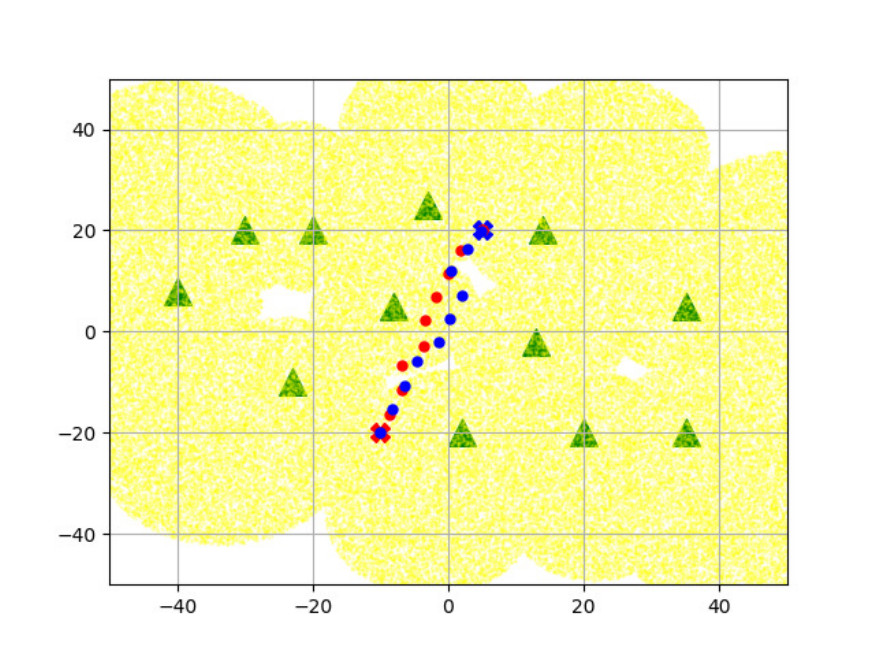}
    		\subcaption{\scriptsize  }
    		\label{Fig:diff_setting_original-2}	
    	\end{minipage}
    	\begin{minipage}{0.32\textwidth}
    		\centering
    		\includegraphics[width=1\textwidth]{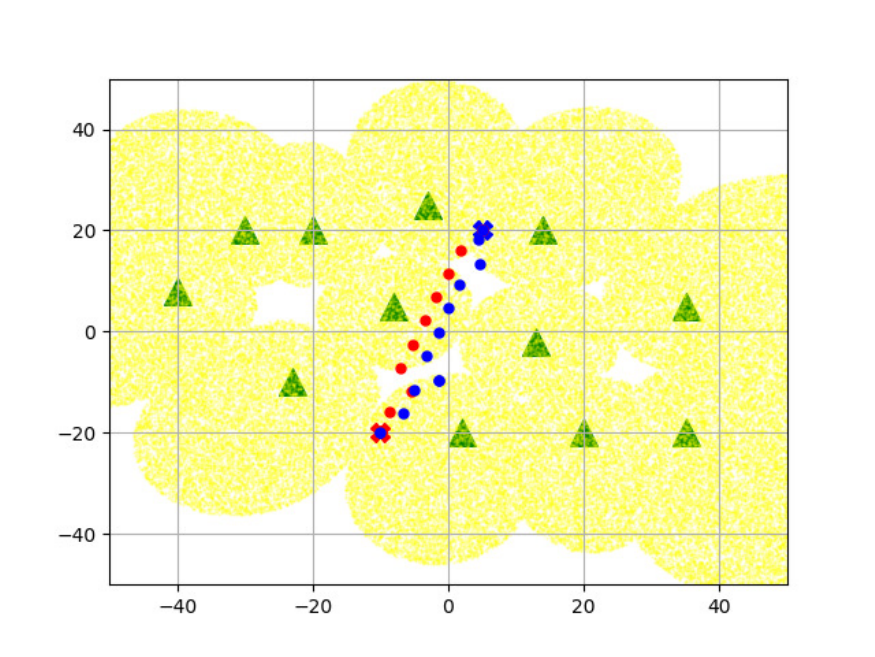}\\
    		\subcaption{\scriptsize }
    		\label{Fig:diff_setting_H100-2}	
    	\end{minipage}
    	\begin{minipage}{0.33\textwidth}
    		\centering
    		\includegraphics[width=1\textwidth]{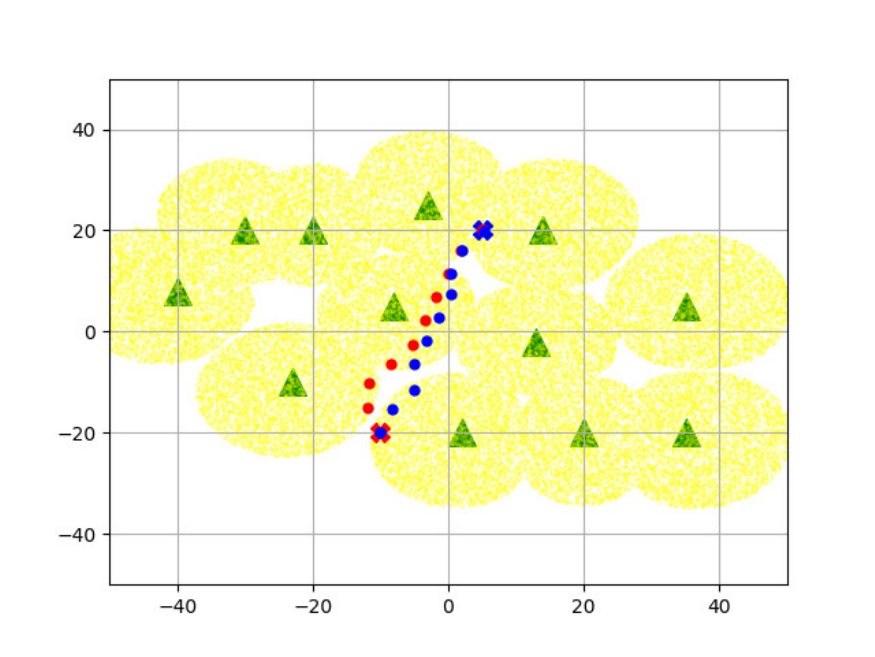}\\
    		\subcaption{\scriptsize }
    		\label{Fig:diff_setting_theta15-2}	
    	\end{minipage}   	
    	\caption{\small Trajectory examples in environments with different settings, i.e., in (a) and (d), $H_V=50$ m, $\theta^{tilt}=10^{\circ}$ and $\theta^{3dB} = 15^{\circ}$; in (b) and (e), $H_V=100$ m, $\theta^{tilt}=10^{\circ}$ and $\theta^{3dB} = 15^{\circ}$;  and in (c) and (f), $H_V=50$ m, $\theta^{tilt}=15^{\circ}$ and $\theta^{3dB} = 35^{\circ}$.  \normalsize}	
    	\label{Fig:diff_setting}
    \end{figure*}

    \subsection{Navigation in Different Settings}
    In this subsection, we present simulation results on the trajectories when the GBSs have different antenna patterns and when the UAVs fly at different heights. The SINR threshold is $\mT_{s} = -4$ dB in this subsection. In. Figs. \ref{Fig:diff_setting} (a) and (d), we provide two different trajectory examples when we have $H_V = 50$ m, $\theta^{tilt} = 10^{\circ}$ and $\theta^{3dB} = 15^{\circ}$. In Figs. \ref{Fig:diff_setting} (b) and (e), UAV altitudes are increased to $H_V = 100$ m, and we notice that due to larger path loss and smaller antenna gains,  coverage zones shrink, which in turn potentially increases the length of the trajectories.  In Figs. \ref{Fig:diff_setting} (c) and (f),   GBSs have larger downtilting angle and 3dB beamwith of the main lobe. In this case, the UAVs experience smaller received power from the main link and potentially larger interference, leading to substantially smaller SINR levels. Therefore, the coverage zones in Figs. \ref{Fig:diff_setting}  (c) and (f) are smaller than those in Figs. \ref{Fig:diff_setting} (a) and (d) and even Figs. 6 (b) and (e).   In all cases, we note that UAVs successfully find different trajectories to meet the connectivity requirements and adapt to different coverage zones.

    \subsection{Navigation in Environments with Obstacles/No-Fly Zones}
    \begin{figure}[h]
    	\centering
    	\begin{minipage}{0.45\textwidth}
    		\centering
    		\includegraphics[width=1\textwidth]{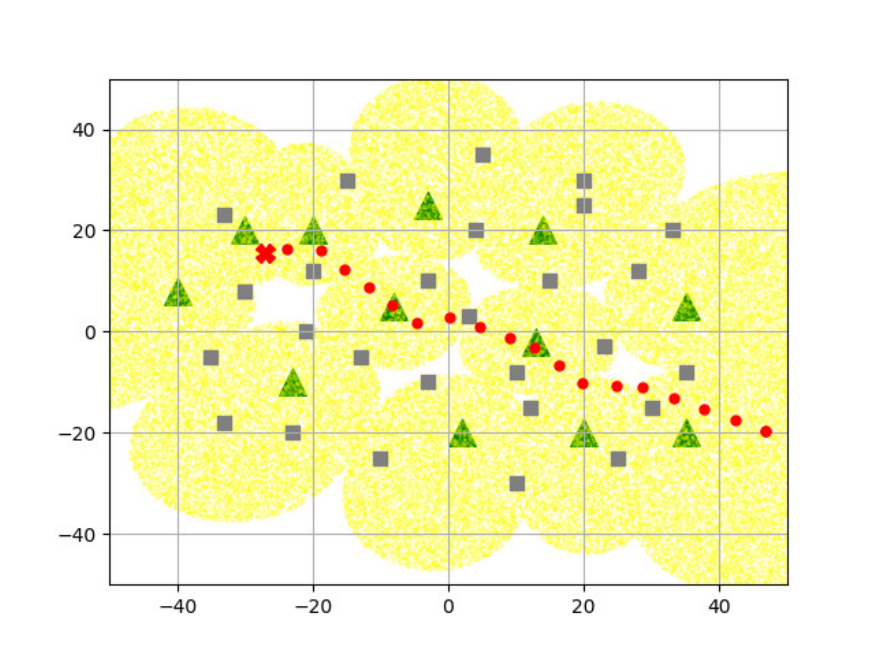}\\
    		\subcaption{\scriptsize Example 1. }
    	\end{minipage}
    	\begin{minipage}{0.45\textwidth}
    		\centering
    		\includegraphics[width=1\textwidth]{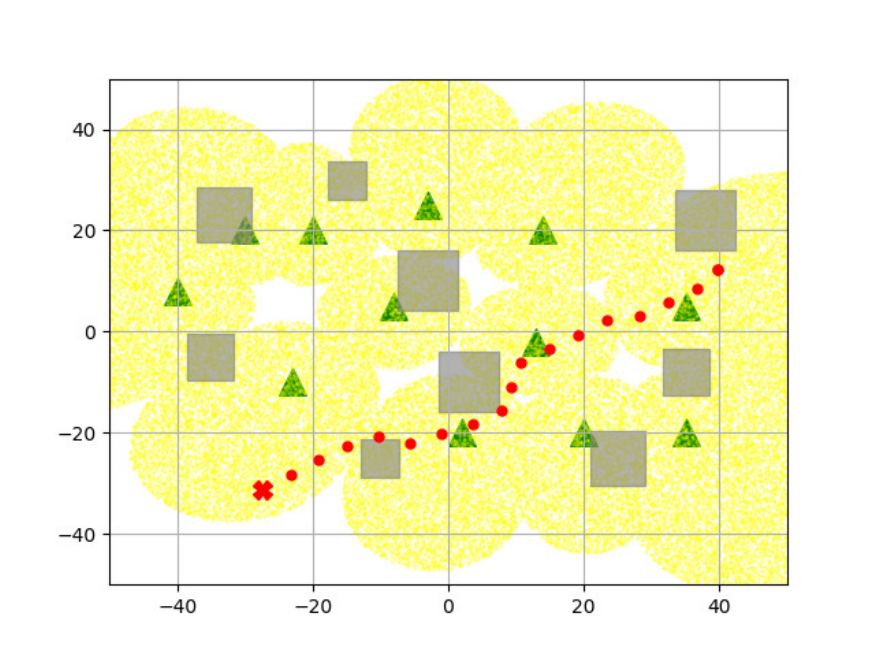}
    		\subcaption{\scriptsize Example 2.}
    	\end{minipage}
    	\caption{\small Trajectory examples in environments with obstacles/no-fly zones.  \normalsize}
    	\label{Fig:obstacles}
    \end{figure}
     The proposed RLTCW-SP algorithm can also be used for navigation in  environment with obstacles that are regarded as non-moving agents. For instance, the trained networks for the 2-agent scenario can be used for  one agent navigation in an environment with obstacles or no-fly zones.  More specifically, the agent can observe the nearest obstacle, and takes the obstacle's location in the joint state for choosing actions. Fig. \ref{Fig:obstacles} displays two illustrations. In this setting, obstacles can be considered as actual obstacles (e.g., tall buildings or structures) or they can model no-fly zones for the UAVs.
     
 \begin{figure}    	
	\centering
	\begin{minipage}{0.45\textwidth}
		\centering
		\includegraphics[width=1\textwidth]{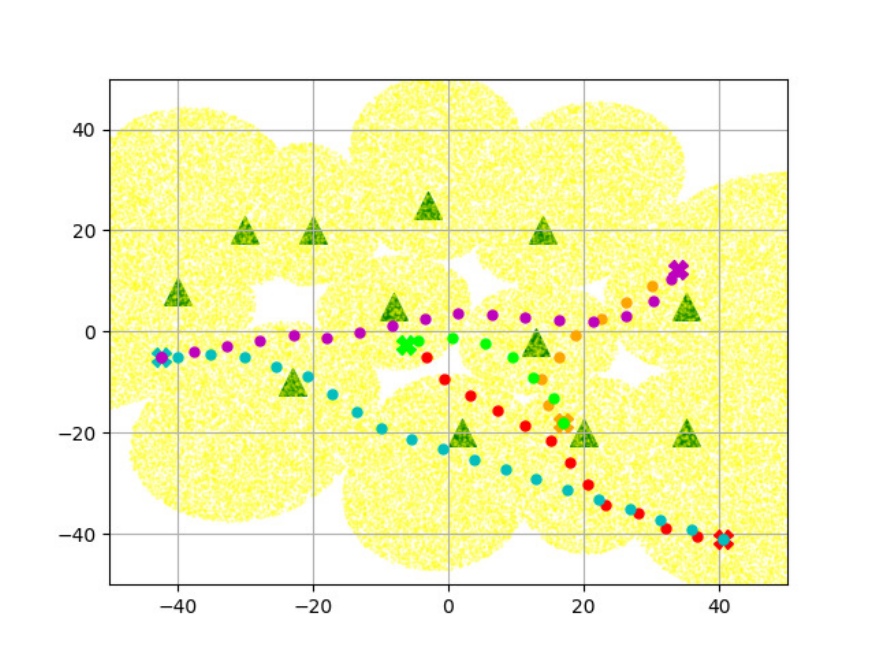}
		\subcaption{\scriptsize Example 1. }
	\end{minipage}
	\begin{minipage}{0.45\textwidth}
		\centering
		\includegraphics[width=1\textwidth]{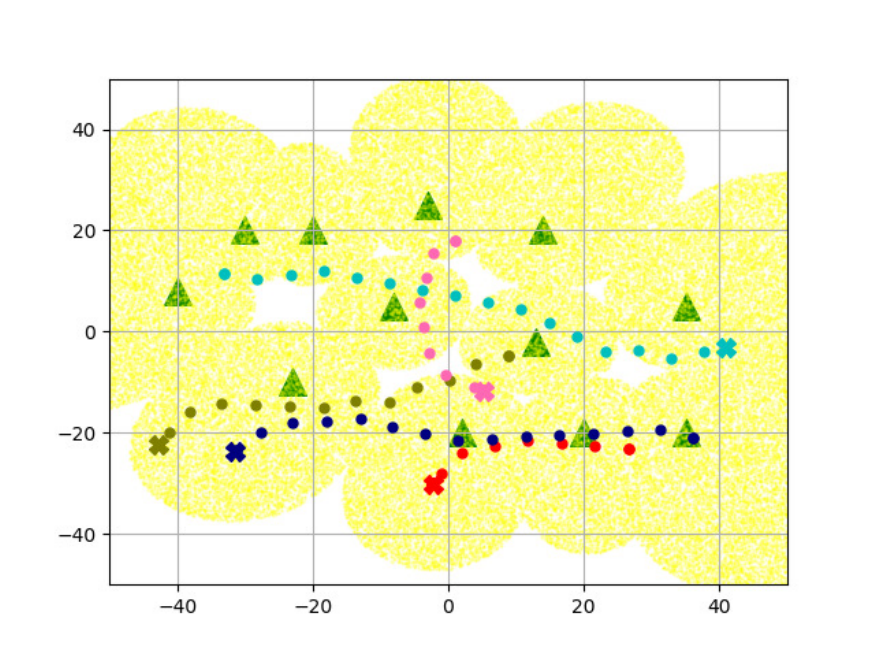}
		\subcaption{\scriptsize Example 2. }
	\end{minipage}
	\caption{\small Trajectory examples for  5-agent navigation.  \normalsize}
	\label{Fig:more_agents}
\end{figure}
\begin{figure}    	
	\centering
	\begin{minipage}{0.45\textwidth}
		\centering
		\includegraphics[width=1\textwidth]{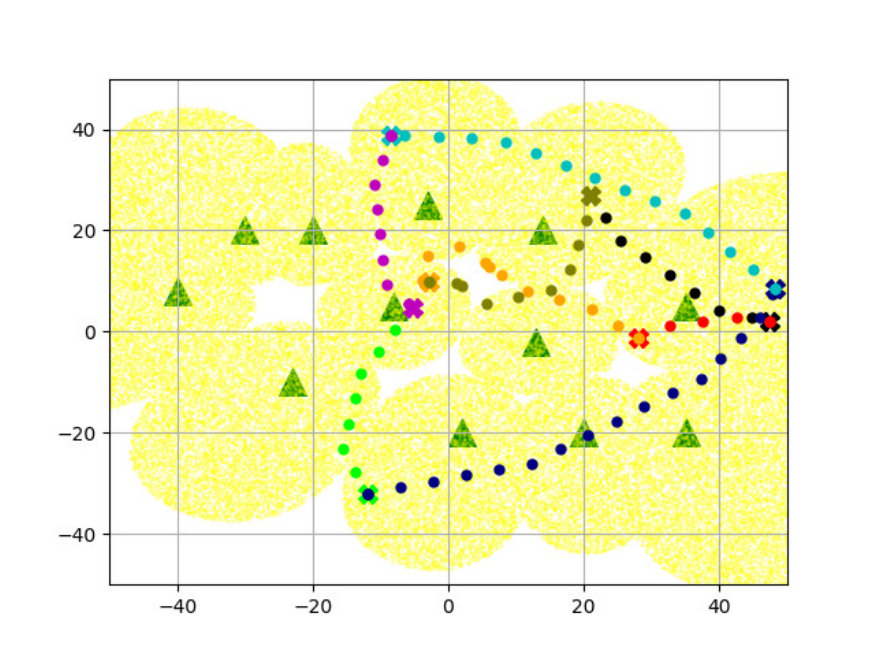}
		\subcaption{\scriptsize Example 1. }
	\end{minipage}
	\begin{minipage}{0.45\textwidth}
		\centering
		\includegraphics[width=1\textwidth]{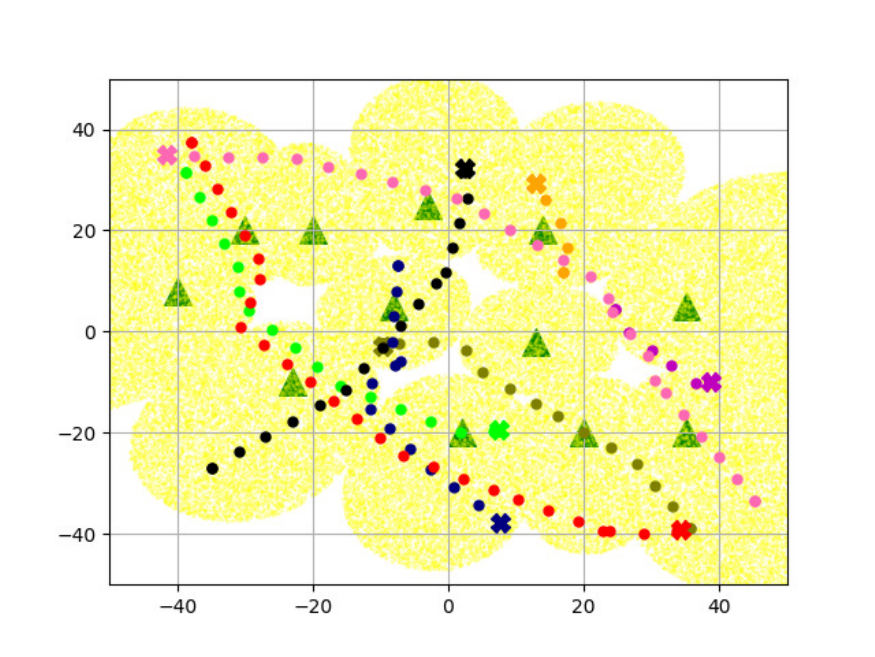}
		\subcaption{\scriptsize Example 2.}
	\end{minipage}
	\caption{\small Trajectory examples for  8-agent navigation. \normalsize}
	\label{Fig:8agents}
\end{figure}

    \subsection{Navigation with More Than Two UAVs}
    Fig. \ref{Fig:more_agents} and Fig. \ref{Fig:8agents} display the illustrations for 5-agent and 8-agent navigation scenarios, respectively.  The SR, CR, DR and AMT performances are presented in Table \ref{Table:more_agents}. We note that the CR increases when more agents are in the environment. As mentioned before, the agents can observe a maximum of 4 nearest agents in the environment. Therefore, for the 8-agent scenario, several agents are non-observable and as a result the CR can increase when compared with scenarios involving smaller number of agents.  On the other hand, when there are more agents in the same area, the interactions become more complex, and harder for the algorithm to handle. However, we note that the performance regarding the SR is still above 90\%. Table \ref{Table:more_agents} also shows  that the agents need more time to reach the destination when there are more agents in the environment.
    \begin{table}[h]    	
    	\centering
    	\caption{Performance for 2/5/8-agent navigation.}
    	\label{Table:more_agents}
    	\begin{tabular}{|c|c|c|c|c|}
    		\hline
    		& SR(\%) & CR(\%) & DR(\%) &AMT(s) \\ \hline
    		2-Agent & 93.02   & 4   & 0.06  & 0.712 \\ \hline
    		5-Agent & 91.12   & 6.32   &  0.8 & 1.001 \\ \hline
    		8-Agent & 90.075   & 7.25   &  0.25  &1.28\\ \hline
    	\end{tabular}
    \end{table}

\section{Conclusion}
In this work, we have studied multi-UAV trajectory optimization with collision avoidance and wireless connectivity constraints. In establishing the wireless connectivity, we have taken into account the antenna radiation patterns, path loss, and SINR levels. We have formulated trajectory optimization as a sequential decision making problem and proposed a decentralized deep reinforcement learning algorithm. In particular, a value neural network has been developed to encode the expected time to the destination given the agent's joint state. An SINR-prediction neural network has been designed, using accumulated SINR measurements obtained when interacting with the cellular network, to map the GBS locations into the SINR levels in order to predict the UAV's SINR levels.  We have investigated the performance in terms of success rate, collision rate, disconnection rate, and average more time. In the numerical results, we have considered various scenarios (e.g., with GBS deployments different from the setting in the training environment, different UAV heights, different antenna patterns, and obstacles/no-fly zones) and  we have shown that with the value network and SINR-prediction network, real-time navigation for multi-UAVs can be efficiently performed in different environments with high success rates.

	
	\bibliographystyle{IEEEtran}
	\bibliography{compresensive2}


\end{document}